\documentclass{article}
\usepackage{iclr2027_conference,times}
\iclrfinalcopy  

\usepackage{amsmath,amsfonts,bm}

\def\eqref#1{equation~\ref{#1}}

\def\1{\bm{1}}

\DeclareMathAlphabet{\mathsfit}{\encodingdefault}{\sfdefault}{m}{sl}
\SetMathAlphabet{\mathsfit}{bold}{\encodingdefault}{\sfdefault}{bx}{n}

\usepackage[hypertexnames=false,hidelinks]{hyperref}
\usepackage{url}
\usepackage{amsmath,amssymb,amsthm}
\usepackage{booktabs}
\usepackage{graphicx}
\usepackage{algorithm}
\usepackage{algpseudocode}
\usepackage{multirow}
\usepackage{array}

\newtheorem{theorem}{Theorem}

\newtheorem{proposition}[theorem]{Proposition}
\theoremstyle{definition}

\theoremstyle{remark}

\newcommand{\mem}{\mathcal{M}}                       
\newcommand{\Amem}{\mathcal{A}_{\mathrm{mem}}}       
\newcommand{\PRMM}{\mathrm{PRM}_{M}}                 
\newcommand{\uRR}{u^{\mathrm{RR}}}                   
\newcommand{\uOC}{u^{\mathrm{OC}}}                   
\newcommand{\uCF}{u^{\mathrm{CF}}}                   
\newcommand{\ind}{\mathbf{1}}                        
\newcommand{\topk}{\mathrm{top\text{-}}k}            
\newcommand{\READ}{\textsc{Read}}
\newcommand{\WRITE}{\textsc{Write}}
\newcommand{\NOOP}{\textsc{Noop}}
\newcommand{\MERGE}{\textsc{Merge}}
\newcommand{\Ntot}{N_{\mathrm{tot}}}

\title{Hindsight Memory-PRM: Supervising Memory\\ Management with Auditable Hindsight Credit}

\author{%
\textbf{Haoxuan Jia$^{1,2*}$ \ \ Yang Liu$^{3*}$ \ \ Yingguang Yang$^{4*}$ \ \ Yancheng Chen$^{5}$}\\
\textbf{Chongyang Zhang$^{1}$ \ \ Hao Zheng$^{1}$ \ \ Qian Li$^{6}$ \ \ Yulin Huang$^{3}$ \ \ Jianshen Zhang$^{3}$}\\
\textbf{Yongzhi Qi$^{3}$ \ \ Shang Luo$^{4}$ \ \ Kefu Xu$^{4}$ \ \ Hao Peng$^{7}$ \ \ Junyu Lu$^{8}$}\\
\textbf{Du Cheng$^{9}$ \ \ Philip S. Yu$^{10}$ \ \ Bin Chong$^{4\dagger}$}\\[4pt]
$^{1}$Fullive-AI \ \ $^{2}$Nanyang Technological University \ \ $^{3}$Supply Chain Tech Team Y, JD.com\\
$^{4}$Peking University \ \ $^{5}$University of Chinese Academy of Sciences\\
$^{6}$Beijing University of Posts and Telecommunications \ \ $^{7}$Beihang University\\
$^{8}$Beijing Institute of Technology, Zhuhai \ \ $^{9}$Northeastern University\\
$^{10}$University of Illinois Chicago\\[2pt]
{\small $^{*}$Equal contribution. \ $^{\dagger}$Corresponding author: \texttt{chongbin@pku.edu.cn}}}

\begin{document}

\maketitle

\begin{abstract}
Memory operations of long-horizon LLM agents are difficult to supervise because their downstream utility becomes observable only after later retrieval and answering. Memory trajectories nevertheless provide machine-readable evidence: which entries are retrieved, cited, and necessary for an answer under a controlled deletion test. \textbf{Hindsight Memory-PRM} uses this evidence offline to train an operation-conditioned memory-utility critic and online to assign intervention-calibrated, entry-level \emph{presence credit}. The credit is propagated along version chains as an action-level proxy reward, without per-operation human labels or full-continuation replay for every action. On held-out LoCoMo, a local 8B policy reaches 77.5\% under a fixed shared reader, compared with 65.1\% for its API teacher and 74.7\% for the strongest reproduced external configuration, while using one eighth as many context tokens as that configuration. The method reaches 79.0\% on LongMemEval. Controlled comparisons separate the effect of observational feedback from that of intervention-calibrated credit, and the learned policy organizes memories into multi-version entries; open-loop controls with similar structure do not recover the full improvement.
\end{abstract}

\section{Introduction}
\label{sec:intro}

LLM agents operating over long horizons must actively manage an external memory \citep{packer2023memgpt,chhikara2025mem0}, deciding turn by turn what to write, merge, or skip. These operations strongly affect downstream success, but their value is not observable when they are taken. A write may become the sole basis of an answer thousands of tokens later, while a skip may discard a fact that will later be queried. With a single episode-level outcome and dozens of decisions, every operation receives the same coarse signal. In our setting, nearly half of the rollout groups receive identical exam scores, leaving no within-group gradient for those steps.

Existing work addresses this problem in two main ways. Prompt-driven managers (Mem0 \citep{chhikara2025mem0}, A-Mem \citep{xu2025amem}, MemBuilder \citep{membuilder2026}) select operations from the current content, but their standard objectives do not directly use subsequent retrieval or answer support. RL-trained managers (Memory-R1 \citep{yan2025memoryr1}, Mem-$\alpha$ \citep{wang2025memalpha}) optimize an outcome-level scalar, which does not identify the contribution of each operation. Process reward models alleviate an analogous problem in mathematical reasoning through human step labels \citep{lightman2024lets,uesato2022solving} or Monte-Carlo resampling with a checkable final result \citep{wang2024mathshepherd,setlur2025rewardingprogress,choudhury2025agentprm}. Memory operations lack a direct step-level correctness signal, and naively resampling them would require replaying the remaining stream at a cost that grows with the horizon.

Our starting point is that memory operations leave machine-readable evidence of delayed utility inside the trajectory. A written entry may later be retrieved, and a retrieved entry may be cited in a correctly answered question. These logs provide partial attribution evidence without per-operation human labels or replaying every continuation. We increase the density of this evidence with auto-generated, source-anchored probe questions and controlled deletion tests (\S\ref{sec:observation}); we refer to this setting as \emph{interventional self-supervision}. The approach extends hindsight relabeling \citep{andrychowicz2017her} and return redistribution \citep{arjona2019rudder,harutyunyan2019hca}, but uses logged retrieval, citation, and intervention outcomes rather than a learned contribution analysis.

Hindsight Memory-PRM uses the audit trail at two levels. Offline, retrieval, citation, and counterfactual evidence define ordinal targets for an operation-conditioned memory-utility critic; counterfactual examples supply supervision for otherwise missing cases such as skipped writes and incorrect merges. Online, each probe exam assigns entry-level credit from retrievals, citations, and one controlled deletion-and-reanswer. This measured credit is combined with a coverage penalty and an annealed critic-shaping term in the GRPO advantage \citep{shao2024deepseekmath}. The measured terms do not depend on the critic, the shaping perturbation has a uniform bound that tightens with annealing \citep{singh1994upper}, and suffix-min shaping is insensitive to isolated increases in non-minimal scores.

On LoCoMo \citep{maharana2024locomo} (5 held-out dialogues, 975 questions), a local 8B policy reaches 77.5\%, compared with 65.1\% for its API teacher, 69.7\% for budget-matched Mem0 \citep{chhikara2025mem0}, and 74.7\% for Mem0's official $k{=}200$ configuration using $8\times$ as many context tokens. The controlled reward-design ladder (63.4 $\to$ 70.2 $\to$ 77.5) shows that intervention-calibrated branches add substantial gains beyond denser observational feedback. The policy develops a multi-version organization; open-loop and rule-based controls that use the same interface do not recover the full benefit. The result also holds on LongMemEval \citep{wu2025longmemeval} (79.0\%), while transfer experiments show that the storage schema remains domain-dependent (\S\ref{sec:transfer}).

\textbf{Contributions.} (i) We formulate hindsight relabeling for memory operations, including zero detection under a stated retrieval invariant and controlled deletion tests for source-anchored probes. (ii) We combine an operation-conditioned memory-utility critic with intervention-calibrated presence credit in GRPO, and characterize the shaping bound and suffix-min rule. (iii) On two long-term memory benchmarks, the resulting local 8B manager outperforms its teacher and the reproduced baselines under a shared reader; controlled comparisons isolate the contributions of observational and intervention-based feedback. (iv) We analyze the learned multi-version organization, its sensitivity to chain-credit normalization, and the effect of the storage schema on cross-benchmark transfer.

\section{Preliminaries and Problem Setup}
\label{sec:prelim}

\paragraph{Setup.}
\label{sec:mdp}
\label{sec:store}
At turn $t$ the agent observes $o_t$ and applies $a_t \in \Amem$ to the bank $\mem_t$ (entries with id, content, layer, timestamp, embedding); transitions are deterministic and the task reward $R(\tau) \in [0,1]$ is end-of-episode QA accuracy. $\Amem$ contains $\WRITE(c,\sigma,key)$, $\MERGE(i,c')$ (append to entry $i$'s version chain, old versions retained), and $\NOOP$; $\READ$ is executed deterministically by the retriever. The store has an \emph{event} layer (timestamped events), a \emph{profile} layer (stable attributes), and an automatic \emph{verbatim} layer preserving every turn---each turn triggers one write gate per managed layer (paths and widths in Appendix~\ref{app:hyper}). The lossless substrate positions the problem as learning a retrieval index and compressed organization, not retention under a budget (\S\ref{sec:limitations}). We exclude destructive updates: deletion is irreversible, lets a policy destroy evidence, and breaks temporal questions (16.2\% of LoCoMo); version chains keep both old and new values observable.

\paragraph{The credit-assignment dilemma.}
\label{sec:dilemma}
With $K$ memory operations per trajectory and an additive outcome $R(\tau) = c_0 + \sum_i c_i(\tau)$, each $c_i$ of variance $\sigma^2$ and pairwise uncorrelated, a baselined policy gradient weights every operation by the same $R(\tau)-b$, so
\begin{equation}
\label{eq:snr}
\mathrm{SNR}_i \;=\; \frac{|\mathbb{E}[c_i]|}{\sqrt{\mathrm{Var}(R - b)}} \;\propto\; \frac{1}{\sqrt{K}} ,
\end{equation}
and sample size to fixed precision scales linearly with $K$ under these assumptions (measured mean $K \approx 37$ in our setting). Eq.~(\ref{eq:snr}) is a dimensional estimate under additivity and zero covariance, but it illustrates the credit-assignment burden for memory-RL methods trained only with a sparse outcome \citep{yan2025memoryr1,wang2025memalpha,zhang2025memact}. Our method is designed to reduce this burden rather than to recover the ideal per-action difference reward.

\section{Hindsight Memory-PRM}
\label{sec:method}

\subsection{Overview}
\label{sec:overview}

\begin{figure}[t]
\begin{center}
\includegraphics[width=\linewidth]{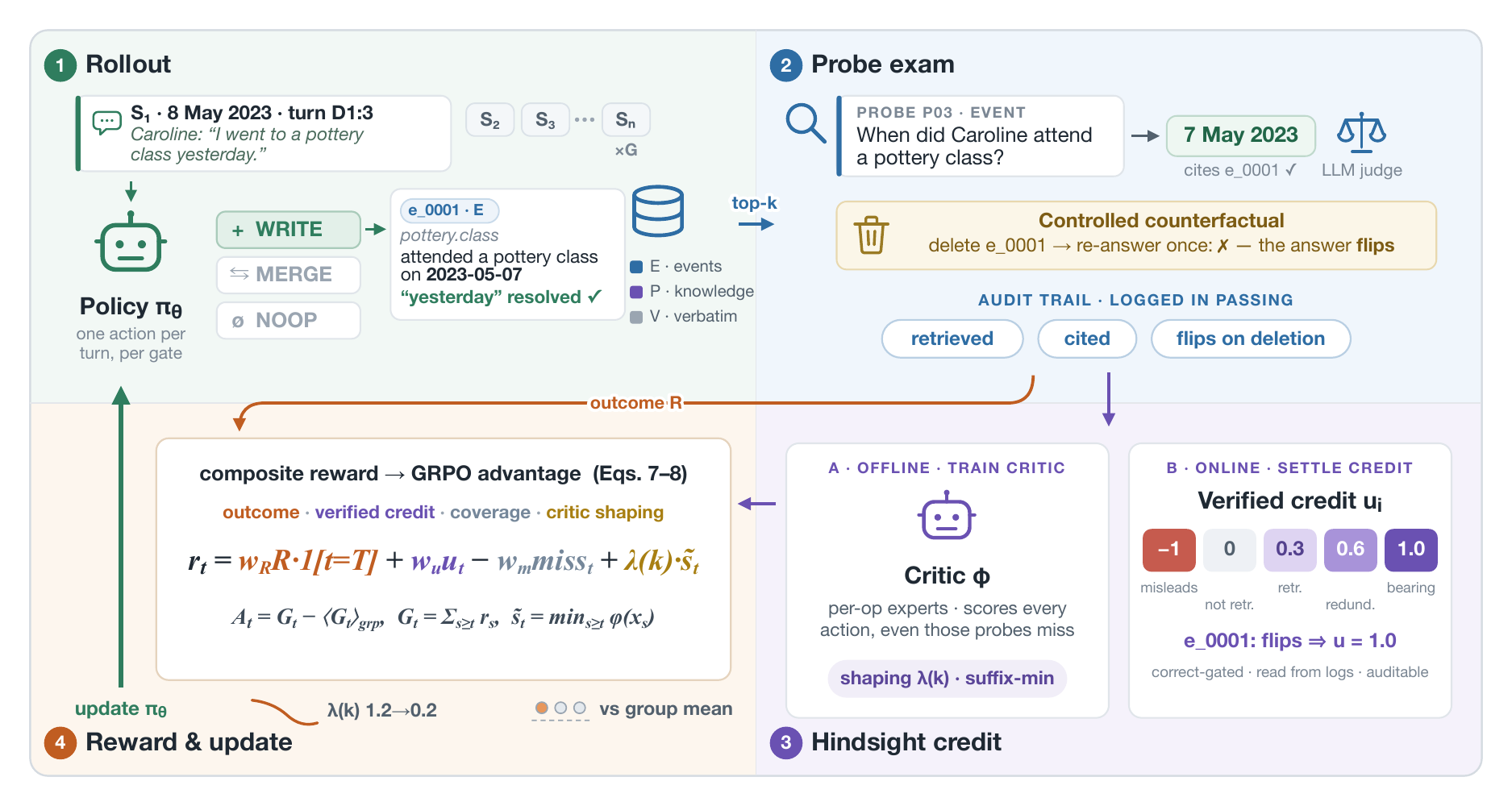}
\end{center}
\caption{Overview. Probe exams retrieve, answer, judge, and run one controlled deletion-and-reanswer per anchored probe; the \textbf{audit trail} (retrieved/cited/flips) trains the critic offline and settles the signed credit of Eq.~(\ref{eq:credit}) online, forming Eq.~(\ref{eq:reward}).}
\label{fig:overview}
\end{figure}

Hindsight Memory-PRM is a loop of three components (Figure~\ref{fig:overview}): the policy rolls out memory management and answering while logging retrievals and citations; a \textbf{hindsight relabeler} converts the logs into per-operation utility targets; the targets train a \textbf{memory-utility critic} $\PRMM$. During optimization the critic supplies an annealed, pessimistic process reward and the relabeler assigns measured credit to each rollout's bank; together with the task outcome, these terms form GRPO's composite advantage. Stage one trains the critic offline on teacher trajectories; it is frozen for the shorter LoCoMo run and periodically refreshed with anchoring and rollback for the longer LongMemEval run (\S\ref{sec:analysis}).\looseness=-1

\subsection{Core observation: memory utility leaves an auditable trail}
\label{sec:observation}

Difference-reward theory \citep{wolpert2001optimal,agogino2008analyzing} credits joint producers not by the total but by
\begin{equation}
\label{eq:diff}
D_i(\tau) \;:=\; R(\tau) - R(\tau^{-e_i}),
\end{equation}
which cancels all randomness not passing through $e_i$ and restores Eq.~(\ref{eq:snr})'s SNR to $O(1)$ (a property of the ideal $D_i$, \emph{not} inherited by our Eq.~(\ref{eq:credit}), \S\ref{sec:analysis}). Measuring it naively takes $O(K)$ replays---but memory operations leave traces making its parts cheap: retrieval logs record recalls, citations record support of correct answers, and both already happened.

\paragraph{Zero detection under the retrieval invariant.} If the pipelines depend on the bank only through retrieval, and entry $e_i$ never enters the candidate set of \emph{any} read path (exam-time $\topk$ or per-layer merge candidates), then every context of $\tau$ and $\tau^{-e_i}$ is token-identical, hence
\begin{equation}
\label{eq:zero}
\uRR_i = 0 \;\Longrightarrow\; R(\tau^{-e_i}) = R(\tau) \;\Longrightarrow\; D_i = 0 .
\end{equation}
Here $\uRR_i=0$ denotes absence from all logged read paths, not merely from the exam top-$k$. The identity rests on one invariant: trajectory generation touches the bank only through retrieval candidates; the critic's bank-wide features are computed post hoc on the reward side and enter no context. On LoCoMo, 74.6\% of entries never enter an exam top-$k$, merge-candidate set, or provenance expansion (deletion-validated in \S\ref{sec:signal}), so Eq.~(\ref{eq:zero}) certifies zero contribution for most writes without replay---a conclusion that, unlike learned return redistribution \citep{arjona2019rudder,harutyunyan2019hca}, follows from the logged execution path under the stated invariant.

\paragraph{Controlled probe construction (one measurement per counterfactual).} On the $\uRR_i = 1$ subset the sign of $D_i$ still needs measuring, and real questions make it infeasible: 52.3\% of LoCoMo evidence turns are shared by several questions, and near-duplicates fill in after deletion, driving the effect size $\delta_i$ below the judging noise $\varepsilon$. Binomial power requires
\begin{equation}
\label{eq:power}
n \;\geq\; \frac{(z_{1-\alpha} + z_{1-\beta})^{2}\,\varepsilon(1-\varepsilon)}{\delta_i^{2}} ,
\end{equation}
with one-sided $\alpha = 0.05$ and power $1-\beta = 0.95$ (i.e., $z_{1-\alpha} = z_{1-\beta} = 1.645$); at $\delta_i \approx 0.039$ and $\varepsilon = 0.05$ this gives $n \approx 340$. We instead construct questions designed to have an effect size close to 1. A candidate probe must pass three gates: it is not answerable from priors, the anchor turn supports the answer, and the remaining turns do not support it. On development re-tests, removing the anchor turn makes 98.7\% of gated probes incorrect (Appendix~\ref{app:gates}). A single deletion-and-reanswer therefore supplies an operational label for the fixed answering model, prompt, and decoding configuration; it is not an information-theoretic guarantee or a statistical proof. Eq.~(\ref{eq:power}) explains why the smaller effect size of real questions would require repeated measurements, while the re-tests in Appendix~\ref{app:gates} assess probe stability. The gates operate on evidence turns, whereas online settlement deletes memory entries. Consequently, a singly anchored probe need not flip after deleting one entry: a non-flip only shows the entry is unnecessary for this fixed reader. The audit attributes most such cases to in-bank redundancy, so Eq.~(\ref{eq:credit}) caps rather than removes their positive credit.

\subsection{Hindsight relabeling: from scalar aggregation to ordinal tiers}
\label{sec:relabel}

A weighted-sum target based on retrieval recall $\uRR$, outcome contribution $\uOC$, and counterfactual gain $\uCF$ performed poorly: across label variants and a two-orders-of-magnitude capacity sweep, development Spearman never exceeded 0.10 (Appendix~\ref{app:calibration}). The low pairwise correlations indicate that these signals capture different properties, making their direct scalar aggregation difficult to fit. We therefore make three revisions. \emph{Citation-calibrated evidence}: the answering model returns the minimal entry-id set it used, which is machine-checked against the retrieved set and logged. Citations receive positive credit only for correctly judged answers; citations on wrong answers enter the toxicity test of Eq.~(\ref{eq:credit}). Calibration shows that citation is a reliable negative filter (100\% recall on entries whose deletion flips an answer, 65\% precision), so it restricts counterfactual measurement to cited candidates. \emph{Content-cluster labels}: a cluster inherits the highest tier of any copy, the earliest writer receives the credit, and later duplicates are penalized (contradictory pairs $-$90\%). \emph{A monotone tier target}: entries are ordered as inert $\prec$ duplicate $\prec$ novel-uncited $\prec$ cited-non-flipping $\prec$ load-bearing. Each boundary corresponds to a logged event or a write-time duplicate test, while the critic conditions only on decision-time information. \NOOP{} is supervised explicitly: because observed misses are rare, every load-bearing write produces a counterfactual same-context \NOOP{} labeled as a miss. Each operation type also receives counterfactual negatives, including well-formed writes on uninformative turns and forced merges where the teacher chose a fresh write. These encourage the critic to use context and estimated utility rather than an operation's surface form.

\subsection{The memory-utility critic}
\label{sec:critic}

The critic scores operations in context: its input concatenates the decision-time context (session prefix, date table, retrieved memories), precomputed bank-state rows (max similarity to existing entries; same-key/topic counts---redundancy as visible input, not something to infer), and the candidate operation. Mixed training degrades the hardest skill (write ranking loses up to 0.2 Spearman), so three per-operation expert adapters---write, merge, \NOOP{}---share one frozen backbone with oracle routing by operation type, reaching each expert's peak simultaneously (\S\ref{sec:validity}). The loss combines utility regression, auxiliary retrievability classification, and in-batch pairwise ranking:
\begin{equation}
\label{eq:loss}
\mathcal{L} \;=\; \mathbb{E}\big[(\varphi(x) - u)^2\big] \;+\; \lambda_{\mathrm{aux}}\,\mathrm{BCE}\big(p(x),\, \uRR\big) \;+\; \lambda_{\mathrm{rank}}\,\mathcal{L}_{\mathrm{rank}},
\end{equation}
with $\mathcal{L}_{\mathrm{rank}}$ a hinge ranking loss over cross-tier pairs (Appendix~\ref{app:losses}). The ranking term is important because tier labels concentrate on adjacent values and squared error alone provides little ranking signal. Adding the term improves development rank correlation sixfold, while the regression term retains a scale anchor for combining shaping with task reward.

\subsection{Online optimization centered on verified credit}
\label{sec:online}

\paragraph{Session-episodes and probe exams.} The online unit is a session window ($\approx$22 turns; $\approx$18--19 gateable, giving $\approx$37 gate decisions per episode). Each episode forks $G$ rollouts from the dialogue's mainline bank, and one final state is written back at random to avoid the optimism of best-of selection. Episode-end exams administer the same quiz to every rollout's bank and yield $\mathrm{acc} \in [0,1]$ as $R$. The continuous outcome distinguishes more rollouts than a binary reward, but ties remain common; zero-variance groups are skipped during optimization. Official questions enter under evidence unlocking, and the final session receives double weight.

\paragraph{Online verified credit.} For every uniquely anchored probe, we remove the cited anchor entry, re-retrieve, and re-answer once. A correct$\to$wrong transition establishes that the entry is necessary for this fixed reader and probe; a non-flip does not distinguish irrelevance from redundancy and therefore receives an intermediate cap. A wrong$\to$correct transition indicates that the cited entry was misleading under the same test. The exam records define a signed ordinal scale:
{\small
\begin{equation}
\label{eq:credit}
u_i \;=\; \ind[\mathrm{correct}_i]\big(0.3 \cdot \ind[\mathrm{retrieved}_i] + 0.3 \cdot \ind[\mathrm{cited}_i] + 0.4 \cdot \ind[\mathrm{flips}_i]\big) \;-\; \ind[\lnot\mathrm{correct}_i]\cdot \ind[\mathrm{cited}_i]\cdot \ind[\mathrm{cures}_i],
\end{equation}
}
where $\mathrm{flips}_i$ ($\mathrm{cures}_i$) indicates correct$\to$wrong (wrong$\to$correct) on re-answer. The positive branch is nested ($\mathrm{flips} \Rightarrow \mathrm{cited} \Rightarrow \mathrm{retrieved}$), giving tiers 0 / 0.3 / 0.6 (cited but not shown necessary by the intervention) / 1.0 (necessary for the fixed reader and probe). The negative branch assigns $-1.0$ when deletion changes a wrong answer to a correct one. The weights set only the tier spacing (sensitivity in Appendix~\ref{app:losses}); unmeasurable positive cases cap at 0.6. Aggregation takes the \emph{highest positive} tier rather than the sum, so serving more probes does not mechanically amplify credit. A measured cure contributes a separate negative term, so positive evidence on other probes cannot mask it.\looseness=-1

\paragraph{Two counterfactuals, two carriers.} Eq.~(\ref{eq:credit}) measures the \emph{presence} counterfactual of the final entry; operations inherit it along version chains, un-normalized and without per-action decomposition (normalized variants and a 300-operation replay validation---Spearman 0.72 against replay marginals---in Appendix~\ref{app:chain}). The action-level counterfactual (``had this not been taken'') would require replaying the continuation, so action-sequence differences are instead supplied by the within-group contrast over identically initialized rollouts. Entry credit refines attribution and reduces variance, although the corrective power of the group contrast is not formally established; the replay sample quantifies the remaining discrepancy. A coverage penalty $\mathrm{miss}_i$ is assigned to empty gates anchored to failed probes. Neither the tier credit nor the coverage penalty depends on a learned scorer. Judge errors and the audit-signal attack surface are examined in \S\ref{sec:validity} and Appendix~\ref{app:erosion}. The additional counterfactual approximately doubles quiz cost.

\paragraph{Composite reward.} The step reward is
\begin{equation}
\label{eq:reward}
r_t \;=\; w_{\mathrm{task}} R\,\ind[t = T] \;+\; w_u u_t \;-\; w_{\mathrm{miss}} \mathrm{miss}_t \;+\; \lambda(k)\,\tilde{s}_t,
\end{equation}
with shaping $\tilde{s}_t$ paid through two branches: memory actions take the suffix minimum after warmup,
\begin{equation}
\label{eq:min}
\tilde{s}_t \;=\; \min_{s \geq t,\; a_s \neq \NOOP} \varphi(x_s), \qquad a_t \neq \NOOP,
\end{equation}
i.e., pessimistic credit (Proposition~\ref{prop:min}); \NOOP{} is paid its expert score $\varphi_{\mathrm{N}}(x_t)$ directly. The asymmetry reflects the different attack surfaces of the two branches. For memory actions, the suffix minimum limits gains from isolated high critic scores. \NOOP{} has no generated content, and excessive \NOOP{} selection is discouraged by counterfactual miss labels, the miss penalty, and its effects on $u$ and $R$. $\lambda(k)$ anneals linearly to $\lambda_\infty$ (Proposition~\ref{prop:anneal}). Each operation's advantage is its undiscounted return-to-go, group same-slot mean-centered \citep{liu2025understanding,yu2025dapo}, and broadcast to its tokens (Algorithm~\ref{alg:training}, Appendix~\ref{app:hyper}). Only $\lambda(k)\tilde{s}_t$ is a learned proxy. Because non-potential shaping can change the optimal policy \citep{ng1999policy}, we treat reward erosion as a risk \citep{gao2023scaling}; annealing reduces the bound and the measured share of the learned term.

\subsection{Analysis of properties}
\label{sec:analysis}

\begin{proposition}[Bounded bias of annealed shaping]
\label{prop:anneal}
With bounded critic outputs and boundedly many shaped steps, the shaped and design objectives differ uniformly by at most $\lambda(k)\,C_B$, giving $2\lambda(k) C_B$-optimal transfer \citep{singh1994upper}---for the \emph{composite design objective}, not the outcome alone; since $\lambda_\infty = 0.2 > 0$, practical force is measured by the shaping share (Remark~2). Proof in Appendix~\ref{app:proof-anneal}.\looseness=-1
\end{proposition}

\begin{proposition}[Pessimism and anti-inflation of min-form credit]
\label{prop:min}
The suffix-min credit never exceeds the current step's score, is 1-Lipschitz, and has zero local sensitivity to non-suffix-minimal scores. Thus, increasing an isolated non-minimal score does not increase credit. The \emph{sum}'s sensitivity to a true bottleneck can still reach the prefix length (Remark~5); the result applies only to the memory branch. Proof in Appendix~\ref{app:proof-min}.\looseness=-1
\end{proposition}

\paragraph{Scope statement.} We do not claim identification of $D_i$: that would require a monotone-evidence assumption contradicted by our redundancy audits (32\% of entry deletions are absorbed by in-bank (near-)duplicates, Appendix~\ref{app:gates}). Eq.~(\ref{eq:credit}) is not an estimator of $D_i$; it identifies the sign on two operational events for the fixed reader (correct$\to$wrong and wrong$\to$correct). Magnitudes are ordinal constants; action-level direction comes from the group contrast and outcome term, not from Eq.~(\ref{eq:diff}).

\paragraph{Critic-update modes.} A frozen critic is a static target, and the over-optimization inflection of agentic process rewards sits at a few hundred steps \citep{choudhury2025agentprm}. Below it we freeze (LoCoMo: 214 steps, 2 epochs); beyond it the critic refreshes every 200 steps on a replay buffer, anchored to the offline snapshot and audited with rollback (LongMemEval; Appendix~\ref{app:erosion}).

\section{Experiments}
\label{sec:experiments}

Two long-term conversational benchmarks answer three questions: does the trained manager surpass existing systems (\S\ref{sec:main}); does the gain come from process reward per se or from how the credit is obtained (\S\ref{sec:signal}); does the learned credit track real memory value (\S\ref{sec:validity})? \S\ref{sec:ablations} attributes components, \S\ref{sec:transfer} covers the second benchmark, transfer, and sample efficiency, \S\ref{sec:emergent} the learned organization.

\subsection{Setup}
\label{sec:setup}

\textbf{Benchmarks.} LoCoMo \citep{maharana2024locomo}: dialogue-level 5/5 split---D1--D4 train, D5 development (calibration/selection halves), D6--D10 (975 questions) test-only, avoiding question-level bank leakage. LongMemEval \citep{wu2025longmemeval}: MemBuilder's protocol---100 training dialogues (50 SFT/40 RL/10 dev), 400 test. Development sets serve checkpoint selection and all design-side checks; test sets serve only final evaluation.
\textbf{Protocol.} All auxiliary roles use one DeepSeek-V4-Flash checkpoint (temperature 0, version pinned, responses cached); all scaffold variants share one BGE retriever \citep{xiao2024bge} with top-10 reading; gate verification uses the exam-answering model, as the operational reading of deletion flips requires. Within our scaffold the manager is the only variable; external frameworks retain their official pipelines, so those comparisons are system-level. Judge--human agreement is 97.5\% on 200 development questions. The policy is Qwen3-8B, the critic Qwen3-1.7B, both LoRA-adapted \citep{qwen2025,hu2022lora}; per-version embeddings, whole-chain return, and the automatic verbatim layer are shared by all our arms.
\textbf{Baselines.} Retrieval-only, prompt frameworks, trained managers (Table~\ref{tab:main}), and two controls isolating our design: an \textbf{MC-rollout critic} reproducing AgentPRM's target \citep{choudhury2025agentprm}, and an \textbf{observational-attribution arm} identical to ours except the online flip/toxicity terms (matrix in Appendix~\ref{app:setup}), so Figure~\ref{fig:ladder}a's last step is attributable to the intervention-calibrated branches alone.
\textbf{Horizon and statistics.} LoCoMo: 214 steps, frozen critic; LongMemEval: 2{,}400 steps, refresh mode; dev-best checkpoints, 3 seeds. LoCoMo tests treat the dialogue as the unit (5 clusters): per-dialogue results, block permutation, sign test $p \approx 0.031$---the 5-unit floor (Appendix~\ref{app:stats}).

\subsection{Main results}
\label{sec:main}

\begin{table}[t]
\caption{LoCoMo held-out test (5 dialogues, 975 questions). Answerer and judge are fixed; external comparisons are \emph{system-level}, ``local 8B'' denotes the manager. Retrieval-only results (46.8--53.4), further prompt-based systems, and the cited Memory-R1 result (62.7): Appendix~\ref{app:setup}, Table~\ref{tab:mainfull}. $\pm$: 3-seed standard deviation; dialogue-level uncertainty: Appendix~\ref{app:stats}.}
\label{tab:main}
\begin{center}
\scriptsize\setlength{\tabcolsep}{4pt}
\begin{tabular}{llrrr}
\toprule
Group & Method & $k$ & Context (tok) & Acc (\%) \\
\midrule
\multirow{2}{*}{System-level reference}
 & Heuristic teacher (our protocol) & 10 & 1{,}320 & 65.1 \\
 & Mem0 (top-10 / official $k{=}200$, 19.6k tok) & 10 & 1{,}150 & 69.7 / 74.7 \\
\midrule
\multirow{6}{*}{Same-scaffold controlled}
 & Untrained (same scaffold) & 10 & 1{,}180 & 58.6 $\pm$ 0.5 \\
 & SFT & 10 & 1{,}640 & 62.5 $\pm$ 0.6 \\
 & Outcome-only GRPO & 10 & 1{,}720 & 63.4 $\pm$ 0.7 \\
 & MC-rollout critic & 10 & 1{,}900 & 65.0 $\pm$ 0.9 \\
 & Observational attribution (online) & 10 & 2{,}210 & 70.2 $\pm$ 0.8 \\
 & \textbf{Ours} (token-matched $k{=}4$: 74.9 $\pm$ 0.9) & 10 & 2{,}480 & \textbf{77.5 $\pm$ 0.8} \\
\bottomrule
\end{tabular}
\end{center}
\end{table}

\paragraph{Overall and budget.} Our method reaches 77.5\%: +18.9 over the untrained backbone, +15.0 over the SFT initialization, and +12.4 over the demonstration teacher, whose performance is therefore not an upper bound. With 2{,}480 context tokens---one eighth of Mem0's official operating point (74.7\% at 19{,}600)---it still leads, on all 5 test dialogues (exact one-sided sign test $p=0.031$, Appendix~\ref{app:stats}); at the token-matched $k{=}4$ setting it leads Mem0 top-10 by 5.2 points. Retrieval-only baselines remain below the untrained manager despite more raw context, answering only 11 of 154 multi-hop questions: organization, rather than raw volume, is the binding factor, consistent with degraded mid-context evidence use \citep{liu2024lost}.

\paragraph{Audit supervision versus MC targets.} The MC-rollout-target critic reaches 65.0\% (+1.6 over outcome-only) at 6$\times$ the API cost of constructing our critic targets. The full audit-based method reaches 77.5\% (+14.1 over outcome-only), although this difference includes both the offline critic and the online audited terms. The controlled ladder in the next section isolates the online intervention-calibrated branches. These results are consistent with memory utility being hard to estimate from sampled continuations and easier to supervise from retrieval, citation, and intervention records.\looseness=-1

\subsection{Construction of the credit signal}
\label{sec:signal}

\begin{figure}[t]
\begin{center}
\vspace{-7pt}
\includegraphics[width=\linewidth]{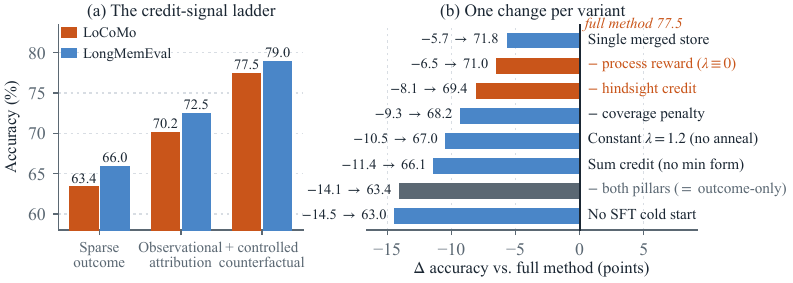}
\end{center}
\vspace{-12pt}
\caption{(a) The credit-signal ladder. (b) Ablations, one change per variant. Orange rows: the two $2\times 2$ pillars ($-$6.5 / $-$8.1 alone; $-$14.1 both removed, gray row), near-additive, interaction $+$0.5; full table in Appendix~\ref{app:setup}.}
\vspace{-4pt}
\label{fig:ladder}
\end{figure}

\paragraph{The signal ladder.} Observational signals are upward-biased when bank redundancy makes a cited entry unnecessary for the answer. Moving from sparse outcome feedback to observational attribution adds 6.8 points; the intervention-calibrated branches add a further 7.3 (Figure~\ref{fig:ladder}a). The last two arms differ only in Eq.~(\ref{eq:credit})'s flip promotion and toxicity term, so this comparison isolates the intervention-based information within our scaffold. Four controls address reward-scale alternatives (Appendix~\ref{app:setup}): matched-scale and shuffled-flip arms reach 70.6 and 70.5; flip-only and toxicity-only reach 75.9 and 71.8. Per episode, roughly 15\% of memory operations receive intervention verification and 26\% tier credit; the rest, including all \NOOP{}s, rely on the critic as their only dense action-local positive signal. The natural-question ablation (Appendix~\ref{app:gates}) confirms this coverage requires constructed probes.\looseness=-1

\paragraph{Validation of zero detection and controlled probes.} Deleting the 74.6\% of entries outside every logged read path changes accuracy by $-0.1$ points; the 500 corresponding deletion pairs are token-identical, so the residual difference reflects provider-side judge noise. Deleting the 7.2\% cited entries reduces accuracy by 12.4 points (Table~\ref{tab:deletion}), limiting intervention measurement to roughly one quarter of the bank. The three gates raise the anchor-removal error rate from 5.2\% on real questions to 98.7\% on constructed probes and reduce the post-removal correct rate from 28.5\% with one gate to 1.3\% with all three (Table~\ref{tab:gates}). At the entry level, 68\% of online deletions flip the answer; the remaining 32\% are not shown necessary for the fixed reader, with the audit attributing most cases to duplicate or near-duplicate evidence (Appendix~\ref{app:gates}).

\subsection{Validity of the critic}
\label{sec:validity}

Removing the critic costs 6.5 points, but end-to-end accuracy alone does not establish validity. All six pre-specified checks on development dialogue D5 pass (Appendix~\ref{app:acceptance}): ranking 0.639, miss AUC 0.951, mis-attach AUC 0.985, cross-expert accuracy 89\%, and adversarial false-positive rate 3.2\%. The most direct check is deletion causality---deleting top-scored entries costs 9.6 points vs.\ 2.1 for random and 0.8 for bottom-scored---and it previously rejected a candidate whose rank correlation passed: correlation measures agreement with the labeling rule, deletion whether that rule tracks downstream value.\looseness=-1

\paragraph{Reward erosion.} Counterfactual negatives, the suffix-min form, and annealing (shaping share 1.8\% late in training) jointly limit reward hacking---content changes also move the measured $u$ and $R$---and we observe no score--performance divergence on LoCoMo. On LongMemEval the critic is refreshed every 200 steps with anchoring, audit, and rollback; five manipulation controls all perform below the no-attack baseline (Appendix~\ref{app:erosion}).\looseness=-1

\subsection{Ablations}
\label{sec:ablations}

\paragraph{The two components form a complete $2\times 2$ factorial} (Figure~\ref{fig:ladder}b). Their simple effects relative to removing both are 6.0 and 7.6 points, with an interaction of $+0.5$; removing either component from the full method costs 6.5 or 8.1 points. The effects are therefore approximately additive. Their scopes also differ: the audited terms $u_t$ and $\mathrm{miss}_t$ apply to probe-touched operations, whereas approximately 30 of 37 actions are \NOOP{}s, for which the critic's direct-payout branch is the only dense action-local positive signal. In this design, the critic primarily supplies coverage and the audited terms supply measured attribution.\looseness=-1

\paragraph{Capacity, initialization, and a negative result.} A 0.6B critic loses 2.9 points; sharing the 8B policy backbone matches the 1.7B critic's accuracy but triples adversarial false positives, motivating the separate critic. Without SFT initialization training still reaches 63.0\%. Pinning high-dependency memories yields no measurable improvement under a strong retriever (details in Appendix~\ref{app:setup}).

\subsection{Second benchmark, cross-benchmark transfer, and sample efficiency}
\label{sec:transfer}

\paragraph{Second benchmark.} The pattern also appears on LongMemEval (Table~\ref{tab:lme}, Appendix~\ref{app:setup}): +18.0 over the untrained backbone and +13.0 over MemBuilder; SFT and ours both produce 76 entries yet differ by 15 points, so entry count alone does not explain the gain. The ladder reads 66.0/72.5/79.0 on 400 independent units (Figure~\ref{fig:ladder}a); the frozen-critic ablation reaches 76.2, 2.8 below refresh mode.\looseness=-1

\paragraph{Transfer: diagnosis and factorization.} Applying the LoCoMo checkpoint directly to LongMemEval reduces accuracy from 77.5 to 58.4. The question distributions require different storage schemas (11 of the 100 LongMemEval training questions concern assistant output, 16 overwritten attributes---neither fits the LoCoMo schema). Rebuilding the bank with a schema designed from LongMemEval training statistics, without retraining the policy, closes 62\% of the gap to native LongMemEval training (Table~\ref{tab:factor}); native retraining accounts for the remainder. The audited policy retains 90\% of native accuracy under the adapted schema, compared with 67\% for outcome-only training. Verified credit thus improves transfer but still requires a domain-appropriate schema.

\begin{table}[t]
\caption{Policy $\times$ schema factorization on the LongMemEval test split; parentheses: accuracy relative to native training (79.0). ``No retraining'' rebuilds the bank under the indicated schema without updating the policy (Appendix~\ref{app:schema}).}
\label{tab:factor}
\begin{center}
\small
\begin{tabular}{lrr}
\toprule
Policy $\backslash$ schema & LoCoMo schema & LME-adapted schema \\
\midrule
Ours (LoCoMo-trained, no retraining) & 58.4 (74\%) & 71.2 (90\%) \\
Outcome-only (LoCoMo-trained, no retraining) & 44.1 (56\%) & 52.6 (67\%) \\
Ours (LongMemEval native) & 63.5 (80\%) & \textbf{79.0} \\
\bottomrule
\end{tabular}
\end{center}
\end{table}

\paragraph{Sample efficiency.} From the same SFT initialization (development accuracy 63.1; Figure~\ref{fig:sample}), outcome-only training uses 28{,}000 rollouts to reach 6\% of the gain to our plateau, while our method passes that plateau after roughly 120 rollouts and finishes at 1{,}712---consistent with, though not predicted by, Eq.~(\ref{eq:snr}). Appendix~\ref{app:stats} reports call-aligned ($\approx$1:50) and end-to-end ($\approx$2.4$\times$) accounting.

\subsection{Learned memory organization}
\label{sec:emergent}

The policy concentrates on merging same-topic facts into multi-version entries (6.1 versions per entry vs.\ the teacher's 1.4, $\pm$0.3 over 3 seeds; Figure~\ref{fig:actions}). Un-normalized chain credit encourages consolidation (Appendix~\ref{app:chain}), so the organization is not claimed to be reward-neutral; however, normalization and cited-version-only controls preserve both structure (5.4/5.2 versions) and accuracy, so the policy learns when and how to use an interface-provided structure rather than creating it.

The mechanism operates through retrieval: every version participates in embedding matching, so one entry supports answers to multiple probes (hit rate 0.57 vs.\ 0.31) while the entry count falls from 412 to 198. This reduces competition among near-duplicate entries. Rule-based merging under the same interface improves slightly but does not recover the full gain (Appendix~\ref{app:chain}): a high merge rate alone is insufficient---the signal must also decide which facts to merge.

\subsection{Limitations}
\label{sec:limitations}

The schema must be designed per domain (\S\ref{sec:transfer}); making it learnable is the key follow-up. Forgetting is under-stressed: LoCoMo has no overturned facts or budget pressure, and the lossless substrate needs extra machinery under deletion rights, calling for a streaming benchmark. Probe generation binds to one closed-source model (mitigated in Appendices~\ref{app:gates}, \ref{app:chain}). Eq.~(\ref{eq:credit}) is conservative---unmeasurable entries never promote and chain-shared credit overrates compensable operations (Appendix~\ref{app:chain})---and acceptance rests on one development dialogue, an external inflection criterion, and relabeling without a convergence guarantee.\looseness=-1

\section{Related Work}
\label{sec:related}

Retrieval-augmented \citep{lewis2020rag,karpukhin2020dpr}, OS-style \citep{packer2023memgpt}, agentic \citep{chhikara2025mem0,xu2025amem}, and reflective \citep{shinn2023reflexion} memory systems differ in construction and retrieval, but in the closest prompt-based managers, later retrieval and answer support do not train each operation. Trained managers use outcome scalars \citep{yan2025memoryr1,wang2025memalpha,zhang2025memact} or observationally attributed rewards \citep{membuilder2026,li2026attrimem}; our controlled comparison adds intervention-based calibration to the latter (\S\ref{sec:signal}). PRMs for mathematical reasoning rely on human step labels \citep{lightman2024lets,uesato2022solving} or Monte-Carlo resampling with checkable answers \citep{wang2024mathshepherd,setlur2025rewardingprogress,choudhury2025agentprm}; memory operations lack such a verifier, and replaying each continuation is expensive. HER, RUDDER, and hindsight credit assignment \citep{andrychowicz2017her,arjona2019rudder,harutyunyan2019hca} redistribute delayed returns; we instead use logged evidence with controlled interventions, motivated by difference rewards \citep{wolpert2001optimal,agogino2008analyzing}; because learned shaping can bias and invite over-optimization \citep{ng1999policy,singh1994upper,gao2023scaling}, we anneal its weight and track its reward share.

\section{Conclusion}
\label{sec:conclusion}

Hindsight Memory-PRM uses retrieval, citation, and controlled-deletion records to train a memory-utility critic offline and to assign intervention-calibrated presence credit online. Across two benchmarks, intervention-based credit improves on denser observational attribution, and the learned manager develops a multi-version memory organization. Extending auditable hindsight supervision to other delayed agent actions, such as caching or planning, is a direction for future work.

\bibliography{iclr2027_conference}
\bibliographystyle{iclr2027_conference}

\appendix
\renewcommand{\theequation}{\Alph{section}.\arabic{equation}}
\setcounter{equation}{0}

\section{Proof of Proposition~\ref{prop:anneal}}
\label{app:proof-anneal}
\setcounter{equation}{0}

\paragraph{Notation and assumptions.} Let $\Theta$ be the parameter space and $J(\theta) := \mathbb{E}_{\tau \sim \pi_\theta}[R(\tau)]$ the design objective. The shaped objective at step $k$ is
\begin{equation}
\label{eq:appA-obj}
J_k(\theta) \;:=\; \mathbb{E}_{\tau \sim \pi_\theta}\Big[\,R(\tau) + \lambda(k)\,S(\tau)\Big], \qquad
S(\tau) \;:=\; \sum_{t:\,a_t \neq \NOOP}\tilde{s}_t \;+\; \sum_{t:\,a_t = \NOOP}\varphi_{\mathrm{N}}(x_t) .
\end{equation}
For ease of statement, the verified credit and coverage penalty of Eq.~(\ref{eq:reward}) are folded into $R(\tau)$: like the outcome they are non-learned, measured signals and part of the design objective; the shaping term refers specifically to the critic's $\lambda(k)\tilde{s}_t$. Two assumptions:
\begin{itemize}
\item \textbf{(A1) Bounded critic outputs.} $\varphi(\cdot) \in [0, B]$; ours passes through a sigmoid, so $B = 1$.
\item \textbf{(A2) Bounded shaped steps.} The number of shaped action steps per episode is at most $\Ntot$. Session-episodes have $\approx$22 turns with two write gates each producing one action per turn, so $\Ntot \leq 44$ (measured mean $\approx$37, \S\ref{sec:ablations}).
\end{itemize}

\paragraph{Step 1: a uniform bound on the objective gap.} Both branches of shaping lie in $[0, B]$: the memory branch $\tilde{s}_t = \min_{s \geq t} \varphi(x_s) \in [0, B]$ and the \NOOP{} branch $\varphi_{\mathrm{N}}(x_t) \in [0, B]$. With (A2), every trajectory satisfies $0 \leq S(\tau) \leq B\,\Ntot$. Taking expectations in Eq.~(\ref{eq:appA-obj}), for \emph{all} $\theta \in \Theta$,
\begin{equation}
\label{eq:appA-bound}
\big|\,J_k(\theta) - J(\theta)\,\big| \;=\; \lambda(k)\,\big|\mathbb{E}_{\pi_\theta}[S(\tau)]\big| \;\leq\; \lambda(k)\,B\,\Ntot \;=:\; \beta_k .
\end{equation}
The same bound holds for every $\theta$, which is the uniformity needed in the next step.

\paragraph{Step 2: $2\beta_k$-optimal transfer.} Let $\theta^\star \in \arg\max_\theta J(\theta)$ and $\theta_k^\star \in \arg\max_\theta J_k(\theta)$ (assuming maxima are attained; otherwise use $\epsilon$-maximizers and add $\epsilon$). Decompose
\[
J(\theta^\star) - J(\theta_k^\star) \;=\;
\underbrace{\big[J(\theta^\star) - J_k(\theta^\star)\big]}_{\leq\, \beta_k \text{ by (\ref{eq:appA-bound})}} +
\underbrace{\big[J_k(\theta^\star) - J_k(\theta_k^\star)\big]}_{\leq\, 0 \text{ by optimality of } \theta_k^\star} +
\underbrace{\big[J_k(\theta_k^\star) - J(\theta_k^\star)\big]}_{\leq\, \beta_k \text{ by (\ref{eq:appA-bound})}} .
\]
Summing,
\begin{equation}
\label{eq:appA-final}
J(\theta^\star) - J(\theta_k^\star) \;\leq\; 2\beta_k \;=\; 2\,\lambda(k)\,B\,\Ntot ,
\end{equation}
i.e., the shaped objective's optimum is $2\beta_k$-optimal under the design objective.

\paragraph{Step 3: annealing contracts the bound.} Since $\lambda(k) = (\lambda_0 - \lambda_\infty)\max\{1 - k/K_{\mathrm{decay}},\,0\} + \lambda_\infty$ is non-increasing with $\lim_{k \to \infty}\lambda(k) = \lambda_\infty$, the right side of Eq.~(\ref{eq:appA-final}) contracts monotonically to $2\lambda_\infty B \Ntot$; at $\lambda_\infty = 0$ the bound is zero and $\theta_k^\star$ is optimal for the design objective (we use $\lambda_\infty = 0.2$; Appendix~\ref{app:losses}). \hfill$\square$

\paragraph{Remark 1 (relation to value-function transfer bounds).} The argument instantiates the standard fact that a uniformly $\epsilon$-approximated objective has $2\epsilon$-optimal maximizers, applied to policy-optimization objectives; it is cognate to Singh \& Yee's bound \citep{singh1994upper} for greedy policies under approximate value functions, but the perturbation acts on the \emph{objective functional} rather than the value function, so no $1/(1-\gamma)$ amplification appears.

\paragraph{Remark 2 (nature of the bound).} Eq.~(\ref{eq:appA-final}) is a worst-case bound independent of task and policy. Annealing makes this upper bound non-increasing, and the bound vanishes when $\lambda_\infty = 0$. In our configuration, however, $\lambda_\infty=0.2$ and $2\lambda_\infty B \Ntot = 17.6$ reward units, the same order as the composite reward's per-episode scale ($\approx$50). The bound is therefore not a tight practical constraint; the more informative quantity is the measured shaping share, which falls to 1.8\% of total absolute reward mass late in training (Appendix~\ref{app:erosion}).

\paragraph{Remark 3 (approximate maximization).} Training does not solve the $\arg\max$; if $\hat\theta$ satisfies $J_k(\hat\theta) \geq J_k(\theta_k^\star) - \epsilon_{\mathrm{opt}}$, then Eq.~(\ref{eq:appA-final}) becomes $J(\theta^\star) - J(\hat\theta) \leq 2\beta_k + \epsilon_{\mathrm{opt}}$: optimization error enters additively.

\section{Proof of Proposition~\ref{prop:min}}
\label{app:proof-min}
\setcounter{equation}{0}

This appendix concerns the memory-action branch; the \NOOP{} branch bypasses the suffix minimum and lies outside Proposition~\ref{prop:min} (\S\ref{sec:online}). Fix a trajectory and its memory-step index set $\mathcal{I} = \{t : a_t \neq \NOOP\}$, write $p_s := \varphi(x_s) \in [0, B]$, and define $\tilde{s}_t = \min_{s \in \mathcal{I},\, s \geq t} p_s$ as in Eq.~(\ref{eq:min}). Of the four properties below, (i)--(iii) are the formal content of Proposition~\ref{prop:min} and (iv) is its direct corollary for reward erosion.

\paragraph{(i) Pessimism.} If $t \in \mathcal{I}$ then $t$ belongs to the index set of the minimum, so $\tilde{s}_t \leq p_t$: credit never exceeds the current step's score, and there is no channel by which later high scores lift an earlier step's credit.

\paragraph{(ii) 1-Lipschitz in the critic outputs.} Let $p, p' \in [0,B]^{|\mathcal{I}|}$ be two score vectors, $i \in \arg\min_{s \geq t} p_s$, $j \in \arg\min_{s \geq t} p'_s$. Then
\[
\min_{s \geq t} p_s - \min_{s \geq t} p'_s \;=\; p_i - p'_j \;\leq\; p_j - p'_j \;\leq\; \|p - p'\|_\infty ,
\]
where the first inequality uses $p_i \leq p_j$ ($i$ minimizes $p$). Swapping roles gives the symmetric side, so $|\tilde{s}_t(p) - \tilde{s}_t(p')| \leq \|p - p'\|_\infty$: bounded critic perturbations cause at most same-size credit perturbations, without amplification.

\paragraph{(iii) Zero local sensitivity to non-minimizers.} Let $j \in \mathcal{I}$, $j \geq t$, $j \notin \arg\min_{s \geq t} p_s$, and let the gap be $\Delta_j := p_j - \min_{s \geq t} p_s > 0$. For any perturbation with $|\eta| < \Delta_j$,
\[
\min_{s \geq t}\,(p + \eta\,\mathbf{e}_j)_s \;=\; \min_{s \geq t} p_s ,
\]
so $\tilde{s}_t$ is constant on a $\Delta_j$-neighborhood of $p_j$ and $\partial \tilde{s}_t / \partial p_j = 0$.

\paragraph{(iv) Corollary: isolated non-bottleneck inflation earns nothing.} Suppose the policy inflates only operation $j$'s score from $p_j$ to $p_j + \eta$ ($\eta > 0$). For each $t \leq j$ with $j \notin \arg\min_{s \geq t} p_s$, (iii) gives $\tilde{s}_t$ unchanged whenever $\eta < \Delta_j$; for $t > j$, $j$ is outside the suffix and $\tilde{s}_t$ is unchanged. The shaping total $\sum_t \tilde{s}_t$ is thus entirely unchanged. To increase total credit one must raise the score of an operation that attains the minimum of some suffix---i.e., improve the weakest operation. This closes the ``compensate one low step with one high step'' cheating channel. \hfill$\square$

\paragraph{Remark 4 (contrast with the sum form).} With $\tilde{s}_t = \sum_{s \geq t} p_s$ (the sum-credit arm of the ablation), $\partial \tilde{s}_t / \partial p_j = 1$ for all $j \geq t$: any single-point inflation increases total credit linearly, and (iii)--(iv) fail. The end-to-end difference is 11.4 points (Table~\ref{tab:ablations}). We note we did \emph{not} observe score--performance divergence in the sum arm (\S\ref{sec:validity}), so its inferiority is supported by end-to-end results rather than direct erosion evidence; its shaping was rescaled before training to the min arm's mean magnitude, so Table~\ref{tab:ablations}'s gap contains no reward-scale or position-bias confound.

\paragraph{Remark 5 (single-point vs.\ total sensitivity).} The 1-Lipschitz property of (ii) is per $\tilde{s}_t$. When a score $p_{s^\ast}$ is the suffix minimum of many prefix steps, the shaping \emph{sum}'s sensitivity to it equals that prefix length, at worst $|\mathcal{I}|$: raising the current bottleneck raises many steps' shaping at once. This does not contradict (iii), which excludes raising \emph{non}-bottleneck scores; raising the true bottleneck is exactly the improvement shaping should encourage, while the disguise channel---making the bottleneck merely \emph{look} better---is not defended by this proposition and is priced on the critic side (counterfactual negatives) and the signal side (memory-bank coupling and the share cap, \S\ref{sec:ablations}).

\section{Complete Loss and Reward Definitions}
\label{app:losses}

\paragraph{Critic loss (Eq.~(\ref{eq:loss})).} The ranking term is
\begin{equation}
\label{eq:rank}
\mathcal{L}_{\mathrm{rank}} = \frac{1}{|P|}\sum_{(i,j) \in P} \max\big(0,\; m - \mathrm{sign}(u_i - u_j)\,(\varphi(x_i) - \varphi(x_j))\big),\quad P = \{(i,j) : |u_i - u_j| \geq \delta\}.
\end{equation} The utility regression is mean squared error on $[0,1]$; the auxiliary classifier uses retrievability $\uRR$ as its label with $\lambda_{\mathrm{aux}} = 0.3$; the pairwise ranking term has $\lambda_{\mathrm{rank}} = 1.0$, pairs built in-batch and restricted to cross-tier pairs with label gap $|u_i - u_j| \geq \delta = 0.1$, hinge margin $m = 0.05$. When a batch contains entries weighted by filler-negative samples, the pair weight takes the larger endpoint. Batch size 4, sequence cap 3072 tokens, left-truncated so the action segment sits at the tail.

\paragraph{Per-operation credit (Eq.~(\ref{eq:credit})).} Indicator weights $(0.3, 0.3, 0.4)$ give the four tiers $\{0, 0.3, 0.6, 1.0\}$. The weights only set tier spacing: any values with $0 < a < b < 1$ preserving the nesting order induce the same ranking; we did not tune them. Sensitivity check: retraining under $(0.2, 0.3, 0.5)$, $(1/3, 1/3, 1/3)$, and $(0.25, 0.25, 0.5)$ yields 77.1--77.6. Each difference from the main configuration is smaller than one reported seed standard deviation, suggesting that the ordinal structure matters more than the precise spacing.

\paragraph{Composite reward (Eq.~(\ref{eq:reward})).} $w_{\mathrm{task}} = 5.0$, doubled for the final-session exam; $w_u = 3.0$; $w_{\mathrm{miss}} = 1.5$.

\paragraph{Shaping schedule.} $\lambda(k) = (\lambda_0 - \lambda_\infty)\max(0,\,1 - k/K_{\mathrm{decay}}) + \lambda_\infty$ with $\lambda_0 = 1.2$, $\lambda_\infty = 0.2$, i.e., $\lambda(0) = 1.2$; the experiments use exactly this form (identical to Appendix~\ref{app:proof-anneal}) with $K_{\mathrm{decay}} = 0.7\,K_{\mathrm{total}}$. The min-form credit of Eq.~(\ref{eq:min}) activates at $K_{\mathrm{decay}}/2$; before that the instantaneous critic score is used, lest early short suffixes degenerate the credit to a single point.

\paragraph{Terms not in the final reward.} Budget and diversity penalties carry zero weight in the final configuration; bank redundancy is controlled by the duplicate penalty of \S\ref{sec:relabel} and write-side credit (\S\ref{sec:online}).

\section{Hyperparameters, Algorithm, and Search Ranges}
\label{app:hyper}

\begin{algorithm}[h]
\caption{Hindsight Memory-PRM training}
\label{alg:training}
\begin{algorithmic}[1]
\Require policy $\pi_\theta$ (SFT cold start), memory environment, synthetic probe pool, retriever
\Ensure trained policy $\pi_\theta$, memory-utility critic $\varphi$
\Statex \textbf{Stage 1 --- offline bootstrap}
\State collect teacher trajectories on the training split; log operation streams, retrieval logs, QA results
\For{each memory operation $i$ of each trajectory}
  \If{$i$ is absent from every logged read path} set $D_i \gets 0$ and $u_i \gets 0$ \Comment{Eq.~(\ref{eq:zero})}
  \Else{} set the tier target $u_i$ by citation calibration and content clustering \Comment{\S\ref{sec:relabel}}
  \EndIf
\EndFor
\For{each synthetic probe with a unique anchor mapping}
  \State delete the mapped entry and re-answer once; update the corresponding $\uCF_i$
\EndFor
\State train per-operation expert critic $\varphi$ with Eq.~(\ref{eq:loss})
\State pre-specified acceptance: ranking / miss / mis-attach / cross-expert scale / deletion causality; on failure, revise the target or critic design
\Statex \textbf{Stage 2 --- policy optimization (critic frozen; periodically refreshed in refresh mode)}
\For{$k = 1 \dots K_{\mathrm{total}}$}
  \State fork $G$ rollouts from the mainline bank snapshot; each manages one session
  \State exam at episode end: retrieve $\topk$ $\to$ answer (with citations) $\to$ judge, get $R$
  \For{each uniquely anchored synthetic probe}
    \State remove the anchor entry $\to$ re-retrieve $\to$ re-answer once \Comment{online controlled counterfactual}
  \EndFor
  \State settle $u_i$ and $\mathrm{miss}_i$ from retrieval, citation, flip, and grade records \Comment{Eq.~(\ref{eq:credit})}
  \State assemble per-step rewards by Eq.~(\ref{eq:reward}) (\WRITE/\MERGE{} shaping takes the suffix min of Eq.~(\ref{eq:min}); \NOOP{} takes the instantaneous expert score $\varphi_{\mathrm{N}}$)
  \State compute group-mean-centered advantages (skip zero-variance groups); update $\pi_\theta$
  \State write one final state back to the mainline at random; advance to the next session
  \If{refresh mode \textbf{and} $k \bmod N_{\mathrm{refresh}} = 0$} \Comment{\S\ref{sec:online}}
    \State $\varphi \gets$ continue training on the replay buffer, anchored to $\varphi_0$
    \State roll back $\varphi$ to the last stable checkpoint if development-set ranking degrades
  \EndIf
\EndFor
\end{algorithmic}
\end{algorithm}

\paragraph{Models and adaptation.} Policy Qwen3-8B \citep{qwen2025}, critic Qwen3-1.7B, both LoRA \citep{hu2022lora} with rank 16, scaling 32, dropout 0.05, applied to all attention and feed-forward projections. The critic's three per-operation experts are separate adapters on a shared frozen backbone, each with a $256 \to 1$ regression head; inference hot-swaps by operation type, and only one backbone resides in memory.

\paragraph{Reinforcement learning.} GRPO \citep{shao2024deepseekmath}, group size $G = 8$; advantages are group-mean-centered without standard-deviation division \citep{liu2025understanding} (small-sample std estimates are unstable and amplify noise as a divisor), zero-variance groups skipped \citep{yu2025dapo}. Episodes are session windows; each dialogue keeps one mainline bank, each episode forks $G$ rollouts by deep copy, and one final state is written back at random. LoCoMo trains 214 steps (frozen mode); LongMemEval uses refresh mode per \S\ref{sec:setup} with $N_{\mathrm{refresh}} = 200$.

\paragraph{Retrieval and answering stack.} Embeddings by BGE \citep{xiao2024bge}. Management decisions take the top-4 per storage layer (12 candidates) so the event layer cannot drown the other layers' merge targets; answer-time retrieval takes a cross-layer flat top-10, followed by one provenance-expansion step: entries originating from the same turn as a hit are appended, so slightly more than 10 entries enter the context. The step applies uniformly to all our rows and cancels in same-family comparisons.

\paragraph{Exams.} Per-episode quiz cap 24; final-session exam cap 60 with double weight; identical quizzes within a group. Answering and judging run with bounded concurrency; single-question failures are skipped without discarding the arm.

\paragraph{Checkpoint selection.} Every configuration saves at fixed intervals, scans on the development set, and evaluates the dev-best on test---identically for all arms.

\begin{table}[t]
\caption{Model roles and decoding configurations. Policy rollout sampling uses temperature 1.0 (for within-group diversity); everything else uses temperature 0.}
\label{tab:roles}
\begin{center}
\scriptsize\setlength{\tabcolsep}{4pt}
\begin{tabular}{llll}
\toprule
Role & Model & Decoding & Where \\
\midrule
Policy (trained) & Qwen3-8B + LoRA & rollout $T{=}1.0$; eval greedy & \S\ref{sec:online}, \S\ref{sec:setup} \\
Critic (trained) & Qwen3-1.7B + LoRA & scoring head, deterministic & \S\ref{sec:critic} \\
Cold-start teacher & DeepSeek-V4-Flash & $T{=}0$ & \S\ref{sec:relabel}, Tab.~\ref{tab:main} \\
Fixed answering model & DeepSeek-V4-Flash & $T{=}0$ & \S\ref{sec:online} exams, all evals \\
Judge & DeepSeek-V4-Flash & $T{=}0$ & \S\ref{sec:online}, \S\ref{sec:setup} \\
Probe generator & DeepSeek-V4-Flash & $T{=}0$ & \S\ref{sec:observation}, App.~\ref{app:gates} \\
Three-gate verification & DeepSeek-V4-Flash (= answerer) & $T{=}0$ & App.~\ref{app:gates} \\
Citation calibration & DeepSeek-V4-Flash & $T{=}0$ & \S\ref{sec:relabel} \\
\bottomrule
\end{tabular}
\end{center}
\end{table}

\paragraph{Search ranges.} Grid-searched: $\lambda_0 \in \{1.0, 1.2, 1.5\}$, $G \in \{4, 6, 8\}$, answering $k \in \{5, 10, 20\}$, critic capacity $\in \{0.6\mathrm{B}, 1.7\mathrm{B}\}$. Everything else keeps the offline-stage settings or follows implementation constraints (e.g., batch size bounded by memory), untuned; Eq.~(\ref{eq:credit})'s tier spacing is likewise untuned.

\paragraph{Cost accounting.} Each LoCoMo seed runs 214 steps with group size 8, or 1{,}712 policy rollouts. Across the three reported seeds, probe generation and evaluation produce approximately 41k candidate probes (22k pass the gates) and 1.6M API calls, costing about \$1{,}100. The reported LongMemEval experiment (2{,}400 steps) uses approximately 210k gated probes and 9.8M calls, costing \$6{,}500. Costs use DeepSeek-V4-Flash pricing.

\paragraph{Optimization details: from per-step reward to token advantage.} Each operation's scalar credit is the undiscounted return-to-go $G_t = \sum_{s \geq t} r_s$ over Eq.~(\ref{eq:reward})'s per-step rewards, through which the terminal task term $w_{\mathrm{task}}R$ enters every prefix operation's credit; the advantage is the group same-slot mean-centered value $A_t = G_t - \bar{G}_t^{\,\mathrm{grp}}$, no std division \citep{liu2025understanding}, zero-variance slots skipped \citep{yu2025dapo}. Same slot means same session turn and same gate: same-turn same-gate operations across rollouts are mutual controls, with missing sides contributing the turn's \NOOP{} $G_t$ to the mean. $A_t$ is broadcast to all output tokens of the operation; the token-level objective is the standard KL-regularized GRPO surrogate, unnormalized by output length (length differences approximately cancel through same-slot contrast). The outcome-only baseline degenerates under the same definition to classic GRPO with $r_s$ containing only the terminal term, i.e., all operations in a rollout share $A = w_{\mathrm{task}}(R - \bar{R})$. Learning rate 1e-5 with cosine decay, KL coefficient 0.02, clip 0.2, one update epoch per batch.

\paragraph{Reproducibility release.} With the paper we release: all gated probes and gate intermediate results, cached API requests and responses, the full prompt set, data splits and dialogue ids, random seeds and all evaluation artifacts, plus scripts that replay evaluation from the cache. The cache fully replays all evaluation and reward computation; policy sampling is local, and only re-collecting teacher trajectories or generating new probes would require live API access (\S\ref{sec:setup}).

\section{Extended Experimental Setup}
\label{app:setup}

\paragraph{Capacity and pinning details (\S\ref{sec:ablations}).} The 8B critic head sharing the policy backbone triples adversarial false positives relative to the separate 1.7B critic (11.4\% vs.\ 3.2\%). Pinning high-dependency memories scores 67.2 vs.\ 67.5 for the unpinned full method; retrieval misses account for 13 of 89 errors.

\begin{table}[h]
\caption{LoCoMo held-out test set (5 dialogues, 975 questions). Every system runs its official construction and reading pipeline end to end; construction model, answering model, and judge are fixed across all prompt-based systems. Comparisons with external systems are \emph{system-level}: our protocol's arms share the three-layer store, per-version embeddings, and the automatically written verbatim layer, while external systems use their official storage and retrieval interfaces; controlled conclusions rest on the same-scaffold arms (teacher, SFT, outcome-only, observational, ours), with interface-equalized controls in Appendix~\ref{app:chain}. Local-8B rows use the policy under evaluation as constructor; ``local 8B'' denotes the memory manager---answering and judging are performed by the same fixed API reader as all other arms. $k$ is the number of retrieved entries; since entry lengths differ across systems we also give the actual answering-context tokens, and we report our method at a token budget matched to Mem0. $^\dagger$Cited from the original paper (own answerer and judge; code unreleased), indirect reference only.}
\label{tab:mainfull}
\begin{center}
\scriptsize\setlength{\tabcolsep}{4pt}
\begin{tabular}{llrrr}
\toprule
Manager & Method & $k$ & Context (tok) & Acc (\%) \\
\midrule
\multirow{3}{*}{\shortstack[l]{System-level ref.:\\ retrieval-only}}
 & RAG-Session & 5 sess & 6{,}300 & 46.8 \\
 & RAG-Utterance & 10 & 4{,}100 & 50.5 \\
 & Full-store (verbatim turns) & 10 & 760 & 53.4 \\
\midrule
\multirow{6}{*}{\shortstack[l]{System-level ref.:\\ prompt-based (API LLM)}}
 & A-Mem & 10 & 980 & 47.9 \\
 & MemoryOS & 10 & 1{,}240 & 55.4 \\
 & MemBuilder-prompt & 10 & 1{,}410 & 59.2 \\
 & Heuristic teacher (our protocol) & 10 & 1{,}320 & 65.1 \\
 & Mem0 & 10 & 1{,}150 & 69.7 \\
 & Mem0 (official operating point) & 200 & 19{,}600 & 74.7 \\
\midrule
\multirow{8}{*}{\shortstack[l]{Same-scaffold controlled:\\ local 8B}}
 & Untrained (same scaffold) & 10 & 1{,}180 & 58.6 $\pm$ 0.5 \\
 & SFT & 10 & 1{,}640 & 62.5 $\pm$ 0.6 \\
 & Memory-R1$^\dagger$ & --- & --- & 62.7 \\
 & Outcome-only GRPO & 10 & 1{,}720 & 63.4 $\pm$ 0.7 \\
 & MC-rollout critic & 10 & 1{,}900 & 65.0 $\pm$ 0.9 \\
 & Observational attribution (online) & 10 & 2{,}210 & 70.2 $\pm$ 0.8 \\
 & \textbf{Ours} & 10 & 2{,}480 & \textbf{77.5 $\pm$ 0.8} \\
 & \textbf{Ours (token-matched)} & 4 & 1{,}090 & \textbf{74.9 $\pm$ 0.9} \\
\bottomrule
\end{tabular}
\end{center}
\end{table}

\begin{table}[h]
\caption{LongMemEval (400-question test split under MemBuilder's protocol; identical questions per row; 3-seed means, std $\pm$0.4--0.9). Retrieval-only rows (38.0--52.0) in Appendix~\ref{app:setup}; questions are independent units, so LoCoMo's nesting issue does not arise (Appendix~\ref{app:stats}).}
\label{tab:lme}
\begin{center}
\small
\begin{tabular}{lrrr}
\toprule
Method & Acc (\%) & Context (tok) & Entries/bank \\
\midrule
Untrained (same scaffold) & 61.0 & 1{,}540 & 364 \\
SFT & 64.0 & 980 & 76 \\
Mem0 \citep{chhikara2025mem0} & 64.0 & 1{,}210 & 185 \\
MemBuilder \citep{membuilder2026} bank + unified reader & 66.0 & 1{,}880 & 458 \\
Outcome-only GRPO (same scaffold) & 66.0 & 1{,}060 & 121 \\
Observational attribution (online, same scaffold) & 72.5 & 1{,}030 & 98 \\
Frozen-critic ablation (no refresh) & 76.2 & 1{,}010 & 82 \\
\textbf{Ours} & \textbf{79.0} & \textbf{990} & \textbf{76} \\
\bottomrule
\end{tabular}
\end{center}
\end{table}

On the 400 LongMemEval test questions, retrieval-only readers score from 38.0 (BM25 over raw sessions) to 52.0 (dense retrieval, $k{=}50$). More raw context does not close the gap, consistent with the LoCoMo evidence that memory organization matters in addition to recall (\S\ref{sec:main}).

\paragraph{Splits in full.} In the official LoCoMo partition, D1--D5 contain 1{,}011 non-test questions. We use only D1--D4 (799 questions) for hindsight labeling and policy training. D5 (212 questions) is held out from training and split by session into D5-cal (10 sessions, 110 questions) for critic acceptance, deletion calibration, and mechanism experiments, and D5-sel (9 sessions, 102 questions) for checkpoint and hyperparameter selection. D6--D10 (975 questions) are used only for final evaluation. LongMemEval's 10 development dialogues are held out from its RL portion, and its 400-question test split is evaluated in full. The two benchmarks share no labels or checkpoints.

\paragraph{Evaluation protocol.} Answering ability and reading budget can confound memory comparisons. Across prompt-based external systems, we therefore use the same base construction model, answerer, judge, and service endpoints while retaining each system's official construction and retrieval logic. The local-8B group instead uses the policy under evaluation as the constructor; within this group, all other components are fixed. Every auxiliary role (teacher, answerer, judge, probe generator, gate verification, citation calibration) uses DeepSeek-V4-Flash in non-thinking mode at temperature 0, version deepseek-v4-flash-20260715, called during July--August 2026 with cached requests and responses.\footnote{No public technical report was available at submission time. We pin the version string and call window and release the request--response cache so that all auxiliary-role calls can be replayed offline.} The only sampled decoding is policy rollout generation (Table~\ref{tab:roles}). Gate verification and exam answering share one model so that a deletion flip has a consistent operational interpretation, and deterministic decoding reduces variation in single re-answers. Shared judges, the short-answer task format, 97.5\% judge--human agreement on 200 development questions, and the different-family reader check in Appendix~\ref{app:chain} provide complementary checks on judge dependence.

\paragraph{Component matrix of the observational arm.}

\begin{table}[h]
\caption{Observational-attribution (online) arm vs.\ the full method.}
\label{tab:matrix}
\begin{center}
\scriptsize\setlength{\tabcolsep}{4pt}
\begin{tabular}{lcc}
\toprule
Component & Observational (online) & Full method \\
\midrule
Offline labels (incl.\ counterfactual completion) and critic checkpoint & identical & identical \\
Probe pool, exams, and miss construction & identical & identical \\
Online positive branch & cited capped at 0.6 & four tiers (flip to 1.0) \\
Online negative branch (toxicity) and deletion-reanswer & absent & present \\
\bottomrule
\end{tabular}
\end{center}
\end{table}

The MC-rollout critic arm keeps backbone, architecture, loss, and recipe identical to ours and swaps only the label source for AgentPRM's Monte-Carlo target \citep{choudhury2025agentprm}; its original ALFWorld checkpoint is out of distribution for memory operations, so we reproduce the construction rather than port weights.

\paragraph{Credit-construction controls.} Four arms test whether the observational$\to$full difference can be explained by reward scale rather than by the placement of intervention labels. (i) \emph{Matched-scale observational}: rescaling the observational reward to the full arm's mean, variance, and nonzero rate gives 70.6 $\pm$ 0.8, compared with 70.2 before rescaling. (ii) \emph{Shuffled-flip placebo}: preserving the tier frequencies while permuting which cited entries are promoted to 1.0 gives 70.5 $\pm$ 0.9. Thus, matching the scale or histogram does not reproduce the full arm. (iii) \emph{Flip-only} without the toxicity branch reaches 75.9 $\pm$ 0.8. (iv) \emph{Toxicity-only} without flip promotion reaches 71.8 $\pm$ 0.9. Within these controls, the flip term accounts for most of the additional improvement, and the two intervention branches are partially additive.

\begin{table}[h]
\caption{Component and mechanism ablations (LoCoMo, 975 questions). Each row removes or replaces exactly one component under the same checkpoint-selection protocol; the sum-credit arm's shaping is rescaled to the min arm's mean magnitude (Appendix~\ref{app:proof-min}, Remark~4).}
\label{tab:ablations}
\begin{center}
\scriptsize\setlength{\tabcolsep}{4pt}
\begin{tabular}{llrr}
\toprule
Variant & Hypothesis tested & Acc (\%) & $\Delta$ \\
\midrule
Remove process reward ($\lambda{\equiv}0$) & net contribution of critic shaping & 71.0 & $-6.5$ \\
Remove hindsight credit & net contribution of verified ground truth & 69.4 & $-8.1$ \\
Remove coverage penalty & complementarity of negative signal & 68.2 & $-9.3$ \\
Constant $\lambda{=}1.2$ & annealing (Prop.~\ref{prop:anneal}: direction only) & 67.0 & $-10.5$ \\
Sum credit & pessimistic min credit (not erosion evidence, App.~\ref{app:proof-min}) & 66.1 & $-11.4$ \\
No SFT cold start & can intervention-calibrated credit bootstrap from zero & 63.0 & $-14.5$ \\
Single merged store & necessity of typed storage & 71.8 & $-5.7$ \\
Critic 0.6B / 1.7B / shared 8B & capacity vs.\ coupling & 74.6 / \textbf{77.5} / 77.2 & --- \\
Pinned layer & does pinning high-dependency memory help & 67.2 vs.\ 67.5 & none \\
\bottomrule
\end{tabular}
\end{center}
\end{table}

\paragraph{Horizon and statistics in full.} LoCoMo: one epoch is the 107 sessions of D1--D4; 214 steps are 2 epochs (frozen mode). LongMemEval: the RL side's 40 dialogues hold $\approx$1{,}900 sessions; one epoch is 18$\times$ larger, and 2{,}400 steps ($\approx$1.3 epochs) is $\approx$6$\times$ the reported inflection, hence refresh mode; sessions are shorter (10.3 vs.\ 21.8 turns) with about half the operations per step. The online counterfactual roughly doubles exam cost and adds 30--50\% wall time per step, charged in the sample-efficiency accounting (Appendix~\ref{app:stats}). All arms scan dev checkpoint grids and evaluate the dev-best on test; 3 seeds with reported standard deviations throughout.

\section{The Three Gates for Synthetic Probes}
\label{app:gates}

Synthetic probes exist to make a single counterfactual decidable in one measurement (\S\ref{sec:observation}). A question must satisfy: the answer is derivable from the anchor turn, only from it, and not from priors. We implement these as three per-question gates; failing any discards the question.

\paragraph{Gate construction.} Let the candidate be anchored at turn $t^{\ast}$.

\begin{table}[h]
\caption{The three gates.}
\label{tab:gatedef}
\begin{center}
\scriptsize\setlength{\tabcolsep}{4pt}
\begin{tabular}{llll}
\toprule
Gate & Context provided & Pass condition & Failure excluded \\
\midrule
\textcircled{1} closed book & no dialogue content & must fail or abstain & answerable from priors \\
\textcircled{2} oracle & anchor turn $t^{\ast}$ only & must answer correctly & dud: anchor does not entail answer \\
\textcircled{3} uniqueness & all content minus $t^{\ast}$ & must fail or abstain & redundant second source \\
\bottomrule
\end{tabular}
\end{center}
\end{table}

Passing all three gates provides operational evidence, for the verifying model and decoding configuration, that the anchor supports the answer, the answer is not available from priors, and the remaining turns do not suffice. We then define $\uCF = \ind[\text{wrong after removing } t^{\ast}]$ in this model-relative sense. Adversarial questions take a separate path: the model must abstain with the full content, filtering out candidates that only appear unanswerable.

\begin{table}[h]
\caption{Effect size, replays required, and label purity of counterfactual measurement. Replays estimated by Eq.~(\ref{eq:power}) (one-sided $\alpha{=}0.05$, power 0.95, $\varepsilon{=}0.05$); $\delta$ is the difference in wrong-answer rate before vs.\ after removing the anchor turn (baseline wrong-answer rate 1.3\%); the last column is the share of questions still answered correctly after removal.}
\label{tab:gates}
\begin{center}
\scriptsize\setlength{\tabcolsep}{4pt}
\begin{tabular}{lrrr}
\toprule
Question source & Wrong after removal & Replays & Correct after removal \\
\midrule
Real questions & 5.2\% & $\approx$340 & 94.8\% \\
Synthetic (one gate) & 71.5\% & 1 & 28.5\% \\
\textbf{Synthetic (three gates)} & \textbf{98.7\%} & \textbf{1} & \textbf{1.3\%} \\
\bottomrule
\end{tabular}
\end{center}
\end{table}

\paragraph{Pipeline and yield.} Questions are generated at the end of every training-split session: an extractor identifies candidate fact points, questions are generated, and each candidate is gated. Extractor recall against the official evidence annotations of the training split is 91.9\% (threshold 90\%). The overall gate pass rate is 51--59\%; by type, it is 97\% for adversarial, 72\% for profile, 53\% for factual, and 33\% for aggregation questions. Among online-measured counterfactuals, deleting the mapped memory entry flips the answer in 68\% of cases. In the remaining 32\%, the entry is not shown necessary for the fixed reader, typically because another entry covers the same fact or the retriever assembles the answer from near-duplicates. This distinction reflects two intervention levels: the gates remove evidence turns when constructing questions (the 98.7\% result in Table~\ref{tab:gates}), whereas online settlement removes memory entries. Eq.~(\ref{eq:credit}) assigns non-flipping entries the capped intermediate tier rather than the load-bearing tier.

\paragraph{Model relativity and independent re-verification.} The gates are behavioral tests of the answering model, not information-theoretic proofs. Table~\ref{tab:gates}'s 98.7\% is a re-test of gated questions on development sessions rather than a restatement of the selection condition: selection uses one wrong answer, whereas the re-test uses the majority of three independent re-answers. A model from a different family re-verifies the three gates on 200 random gated probes with 93.5\% agreement. Replacing the temperature-0 re-answer with a majority of three samples at temperature 0.7 gives 97.0\% agreement, and the anchor-removal wrong-answer rate remains above 96\% under two independently rewritten answer prompts. These checks support stability under model, prompt, and decoding changes while preserving the operational, model-relative interpretation.

\paragraph{Deletion protocol.} Synthetic probes are singly anchored by construction; the controlled counterfactual deletes exactly the one entry mapped from the question's anchor evidence; auxiliary entries the answer also cited are not deleted, and multiple citations trigger no joint deletion. The anchor mapping is maintained by write-time source-turn ids: every entry stores the set of turns its content came from, and the mapping can be one-to-many (one turn split into several entries) or many-to-one (several turns merged into one). The deletion target is the entry whose sources contain the anchor turn and which this question cited---unique for 99.1\% of gated probes; the remaining 0.9\% are mapping-ambiguous and treated as counterfactually unmeasurable (positive branch capped at 0.6, negative branch not triggered).

\paragraph{Training-side dependency ablations.} Four ablations test dependence on the probe generator and official-question injection. (1) Generating training probes with a model from a different family, while retaining the fixed answering model for the gates, gives 76.8 $\pm$ 0.9 after retraining, 0.7 points below the main configuration. (2) Removing official-question unlock injection and using only synthetic probes for the outcome gives 76.9 $\pm$ 0.8. Both differences are smaller than one reported seed standard deviation, suggesting that the result is not specific to one generator or to official-question injection. (3) With no synthetic probes, official unlocked questions alone give 72.6 $\pm$ 1.0; intervention-verification coverage falls to 3\%. (4) Retaining synthetic probes but removing source-turn mapping gives 73.9 $\pm$ 0.9 because misses cannot be localized and anchored deletion cannot be executed. The last two ablations show that the constructed exam environment and source anchoring make substantial contributions to usable supervision density.

\paragraph{Relation to official questions.} Synthetic questions produce training signal only; evaluation uses official questions throughout, and test-split official questions take part in no generation, training, or checkpoint selection. Training-split official questions are injected into the exam pool under evidence unlocking (\S\ref{sec:online}), providing an outcome signal distributionally matched to evaluation without joining counterfactual measurement.

\paragraph{Limitations of probe construction.} The aggregation type has a low pass rate (33\%), limited by the generator's cross-session aggregation ability. Post-gate type proportions also deviate from the pre-set quotas; future implementations should define quotas over accepted probes rather than generated candidates.

\section{Construction of the Pre-specified Acceptance}
\label{app:acceptance}

\begin{table}[h]
\caption{Pre-specified critic acceptance (development dialogue D5). Thresholds fixed before the experiment; all checks run on D5, which takes part in neither critic nor policy training; final test D6--D10 takes part in no check.}
\label{tab:acceptance}
\begin{center}
\scriptsize\setlength{\tabcolsep}{4pt}
\begin{tabular}{llll}
\toprule
Check & Measures & Threshold & Result \\
\midrule
Dev ranking & Spearman of scores vs.\ hindsight targets & $\geq 0.55$ & 0.639 \\
Miss detection & true \NOOP{} vs.\ counterfactual miss (AUC) & $\geq 0.85$ & 0.951 \\
Mis-attach detection & true merges vs.\ forced attaches (AUC) & $\geq 0.80$ & 0.985 \\
Cross-expert scale & correct action outscores wrong action & $\geq 75\%$ & 89\% \\
Style adversary & false-positive rate on keyword-stuffed probes & $\leq 5\%$ & 3.2\% \\
Deletion causality & loss deleting top / random / bottom 10\% & $\geq 8$ / mid / $\leq 1$ & $9.6$ / $2.1$ / $0.8$ \\
\bottomrule
\end{tabular}
\end{center}
\end{table}

The six checks and thresholds of Table~\ref{tab:acceptance} were written down before critic training and served as the sole basis for freezing. Development dialogue D5 never takes part in critic training.

\paragraph{\textcircled{1} Dev ranking.} On development dialogue D5, rebuild each E/P entry's (context, action) pair at its creating write, score with the write expert, and compute Spearman correlation against the entry's hindsight tier target. The 0.55 threshold is a usability judgment (matching Table~\ref{tab:acceptance}): below it, adjacent-tier ranking errors within a group would exceed 30\% and the shaping term's contribution to the advantage approaches randomness.

\paragraph{\textcircled{2} Miss detection.} Positives are the teacher's true \NOOP{}s on turns with nothing to record; negatives are counterfactual---a load-bearing write replaced by a same-context \NOOP{} labeled miss. AUC threshold 0.85.

\paragraph{\textcircled{3} Mis-attach detection.} Positives are the teacher's true merges; negatives are counterfactual forced attaches onto nearby entries where the teacher clearly chose a fresh write. Threshold 0.80.

\paragraph{\textcircled{4} Cross-expert scale.} Using the ready-made counterfactual pairs, compare across experts on the same turn: on turns that should be written, the write expert's score for the true write must exceed the \NOOP{} expert's score for the same-turn counterfactual \NOOP{}; on turns with nothing to record, the reverse. This checks the three experts share one scale; threshold 75\% of pairs.

\paragraph{\textcircled{5} Style adversary.} Construct candidate writes that are semantically irrelevant to any question but written in keyword-stuffed high-frequency style, pair them with true high-utility writes, and measure how often the critic scores them high. Threshold: false-positive rate $\leq$5\%. This measures fragility on an adversarial distribution, complementing \textcircled{1}--\textcircled{3} on the natural distribution.

\paragraph{\textcircled{6} Deletion causality.} Build three deletion variants of development dialogue D5's bank (top-scored 10\%, random 10\%, bottom-scored 10\%; equal counts, fixed seed for random) and compare on the same official questions with identical reading and answering components; the only variable is the deleted set. Pass requires the top-deletion loss to significantly exceed random and the bottom-deletion loss to be near zero.

\paragraph{Revision and rollback path.} A failed threshold triggers a review of both the target construction and the critic design rather than an automatic increase in capacity. In one iteration, checks \textcircled{1}--\textcircled{3} passed but deletion causality (\textcircled{6}) failed because deleting top-scored entries reduced accuracy less than deleting random entries. We rejected that candidate and reverted to the previous target-and-critic configuration. Rank correlation measures whether the critic reproduces the labels; deletion causality additionally tests whether those labels track downstream memory value.

\section{Anti-erosion: Implementation and Records}
\label{app:erosion}

\paragraph{The shaping share.} Table~\ref{tab:share} gives the mean contribution of each reward component to the per-action return-to-go $G_t$. Shaping is the only learned proxy of the four; the other terms come from retrieval and citation records, the source turns of failed probes, and judge grades. Over the three reported phases, annealing reduces the shaping share from 9.6\% to 1.8\%. This measurement bounds the learned term's contribution in the reported LoCoMo run, but does not establish horizon-independent protection.

\begin{table}[h]
\caption{Reward composition over training (LoCoMo, full method). Thirds of training steps; parentheses give the share of the four-term absolute sum; the miss share falls as the policy stops missing anchors (cf.\ Figure~\ref{fig:actions}).}
\label{tab:share}
\begin{center}
\scriptsize
\begin{tabular}{lrrrr}
\toprule
Phase & Shaping (forgeable) & Hindsight credit (audited) & Coverage (miss) & Task (measured acc) \\
\midrule
First third & 0.563 (9.6\%) & 1.043 (17.8\%) & 0.420 (7.2\%) & 3.826 (65.4\%) \\
Middle third & 0.218 (4.0\%) & 1.436 (26.6\%) & 0.246 (4.6\%) & 3.497 (64.8\%) \\
Last third & 0.106 (1.8\%) & 2.007 (34.1\%) & 0.118 (2.0\%) & 3.658 (62.1\%) \\
\bottomrule
\end{tabular}
\end{center}
\end{table}

\paragraph{Refresh implementation.} Every $N_{\mathrm{refresh}} = 200$ steps the critic trains one round on replay-buffer samples with the loss of Eq.~(\ref{eq:loss}), applying the anchor $\theta \leftarrow (1-\lambda_{\mathrm{anc}})\theta + \lambda_{\mathrm{anc}}\theta_0$ after each micro-batch, $\theta_0$ the offline snapshot. After refresh, ranking correlation and discrimination AUC are re-checked on the development set; if any metric falls below its preset floor or drops beyond tolerance from the historical best, the critic rolls back to the last stable checkpoint and skips the refresh. Anchoring suppresses drift; audit and rollback provide the stop-loss; without either, refresh itself becomes a new over-optimization channel.

\paragraph{Choosing the interval.} $N_{\mathrm{refresh}}$ obeys two constraints. Safety margin: between refreshes the critic is effectively frozen, and 200 keeps this effective frozen horizon within half the reported inflection \citep{choudhury2025agentprm}. Data volume: LongMemEval sessions average 10.3 turns, each rollout step yields $\approx$19 memory operations, and at group size 8, 200 steps accumulate $\approx$30k hindsight-labeled operations---the offline bootstrap's order, enough for one effective refresh without waiting longer. With the RL side's 40 dialogues and $\approx$1{,}900 sessions, one epoch holds 9--10 refresh points; the reported run is 2{,}400 steps and 12 refresh points, $\approx$1.3 epochs.

\paragraph{Audit record on LongMemEval.} Across 12 refresh points, development ranking correlation rises from 0.62 to 0.66 with no monotone decline; one rollback triggered. The audit metric is a component of the refresh mechanism, not an independent referee---it detects the critic degrading relative to the development distribution but cannot exclude the critic and that distribution drifting together.

\paragraph{Extended attack surface.} Style stuffing tests only the critic, so we also evaluate five manipulations of the audit signal: query copying (writing probe text into entries), citation baiting (embedding instructions to cite an entry), answer templating, oversized entries, and prompt injection directed at judging. With the correctness gate and toxicity tier of Eq.~(\ref{eq:credit}), query copying and oversized entries increase retrieval but reduce answer correctness; citation baiting and prompt injection do not reach the judge because it sees only the question, reference answer, and model answer; and templated answers fail the correctness check. All five variants score 3.1--9.4 points below the no-attack control, providing no evidence that these manipulations improve the composite objective.

\paragraph{Adversarial probe construction.} We construct candidate writes semantically irrelevant to the questions but keyword-stuffed, pair them with true high-utility writes, and measure the critic's high-utility rate (Table~\ref{tab:acceptance}). This tests whether the critic mistakes style for utility, complementing ranking and deletion causality: those measure quality on the natural distribution, this one fragility on an adversarial one.

\section{Construction and Calibration of Hindsight Targets}
\label{app:calibration}

This appendix gives the quantitative basis for \S\ref{sec:relabel}.

\paragraph{Numeric mapping and class composition of the five tiers.} Mapping: inert 0, duplicate 0.15, novel-uncited 0.3, cited-redundant 0.6, load-bearing 1.0; over 20.6k teacher operations the tier shares are 41\% / 12\% / 28\% / 13\% / 6\%. Counterfactual augmentation adds 2.1k miss negatives, 1.9k filler writes, 1.4k mis-attach merges. Regression uses the numeric mapping; ranking uses only the order.

\paragraph{Empirical failure of weighted aggregation.} Across five weighted label variants and a capacity sweep from LoRA to full fine-tuning (a 94$\times$ parameter range), development-dialogue Spearman never exceeds 0.10. On entries with measured counterfactuals, retrievability and counterfactual gain have $\rho = 0.05$; a separate manual importance score and counterfactual gain have $\rho = -0.07$. These low correlations show that the signals need not align, while the label-variant results establish that their direct weighted aggregation performed poorly in our setting.

\begin{table}[h]
\caption{Deletion effects by credit stratum (development calibration segment D5-cal; of its 110 questions, 104 admit unique anchor mapping and controlled deletion---no sampling). Rows are 3-seed means; confidence intervals use session-level block bootstrap over D5-cal's 10 sessions (blocks absorb within-dialogue correlation), 95\% half-widths 0.6--1.4 points.}
\label{tab:deletion}
\begin{center}
\small
\begin{tabular}{lrr}
\toprule
Deleted entry set & Share of bank & $\Delta$ Acc \\
\midrule
Never retrieved ($\uRR{=}0$) & 74.6\% & $-0.1$ \\
Retrieved but uncited & 18.2\% & $-1.9$ \\
Cited ($\uOC{=}1$) & 7.2\% & $-12.4$ \\
\bottomrule
\end{tabular}
\end{center}
\end{table}

\paragraph{Citation-protocol calibration.} On 154 single-target deletions: every entry whose deletion caused a wrong answer appears in the citation set (recall 100\%), with no hallucinated entry ids; reverse precision is 65\%, i.e., about a third of cited entries do not change the result when deleted. Citation is therefore a reliable negative filter, not a positive criterion---which fixes its role in the pipeline: restricting expensive counterfactual measurement to a candidate set and feeding the intermediate tiers of Eq.~(\ref{eq:credit}); citation alone never establishes load-bearing credit---the 1.0 tier requires a measured flip.

\paragraph{Effect of content clustering.} 34\% of teacher writes rephrase existing content. Before clustering, development dialogues contain 145 pairs with cosine $\geq 0.9$ and label gap $\geq 0.4$; after clustering at 0.9 with first-writer attribution and duplicate penalty, 18 pairs remain.

\paragraph{Effect of the ranking term.} On identical labels, switching the loss from pure regression to regression-plus-ranking lifts development-dialogue Spearman from 0.083 to 0.533.

\section{Storage-Schema Adaptation on LongMemEval}
\label{app:schema}

The two benchmarks differ sharply in dialogue structure, and the three-layer schema of \S\ref{sec:store} does not directly match LongMemEval. We retain the three-layer structure but redefine each layer's content and write rules. The change is confined to gate prompts and merge rules; credit construction, the critic architecture, and optimization remain fixed. This factorization therefore measures the effect of schema adaptation within the same training method. The adaptation uses question-type statistics from the 100 training questions; the test split is not used, and all counts below refer to the training split.

\begin{table}[h]
\caption{Question-type distribution (LongMemEval training split).}
\label{tab:lmetypes}
\begin{center}
\small
\begin{tabular}{lrl}
\toprule
Type & Share & Asks about \\
\midrule
multi-session & 27/100 & cross-session aggregation and counting \\
temporal-reasoning & 26/100 & temporal location and arithmetic \\
knowledge-update & 16/100 & attributes overwritten by later statements \\
single-session-user & 14/100 & user-stated attributes \\
single-session-assistant & 11/100 & \textbf{content the assistant previously produced} \\
single-session-preference & 6/100 & answering preferences \\
\bottomrule
\end{tabular}
\end{center}
\end{table}

Five mismatches and their fixes:

\paragraph{(1) Speaker asymmetry.} LoCoMo has two symmetric speakers narrating their lives; LongMemEval has user and assistant, and 11 of the 100 training questions ask directly about the assistant's prior output (we verified their evidence turns carry \texttt{role=assistant}). The event layer therefore accepts assistant-side events, prefixes every record with \texttt{user:} or \texttt{assistant:} to preserve provenance, and appends to assistant events an anchor for which request they answered.

\paragraph{(2) Attributes get overwritten.} LoCoMo's stream is a consistent biography without overturned values; 16 of the 100 training questions probe both sides of an overwrite. The knowledge layer therefore enforces a \texttt{current / since / previous} triple for mutable attributes and requires same-key new values to merge into the original entry rather than open a second one; otherwise retrieval returns both values while questions ask both ``what is it now'' and ``what was it before.''

\paragraph{(3) Subjects beyond people.} The original profile layer organizes ``speaker $\to$ attribute''; much of LongMemEval's stable knowledge concerns objects, places, or user-context concepts. The knowledge layer extends its subject range, explicitly excludes common knowledge the reader already has, and splits by aspect to prevent entry bloat.

\paragraph{(4) Procedural content is not represented.} The assistant often produces tutorials, schedules, and recipes, which do not fit the original layer definitions. The adapted knowledge layer adds a procedure class with the fixed format ``purpose | steps | applicable context,'' retaining quantities, times, and table rows verbatim.

\paragraph{(5) Counting aggregation.} Most of the 27 multi-session training questions count instances of repeatable events. Event keys carry a repeatable-class prefix so same-class events group, and the merge rule becomes \emph{aggregate rather than summarize}: same-class events merge into one span-covering entry that must enumerate every instance with dates and counts.

\paragraph{Correspondence to existing systems.} The adapted three layers map onto MemBuilder's four memory types \citep{membuilder2026}: its Episodic type corresponds to our event layer; its Semantic and Procedural types correspond to the knowledge layer's entity and procedure classes; and its resident Core corresponds to the knowledge layer's state and preference classes. We do not provide Core with a separate resident channel. The LoCoMo ablation in \S\ref{sec:ablations} finds no measurable improvement from pinned copies when the retriever is strong. Whether this result transfers to LongMemEval remains untested, because approximately 20\% of its training questions concern user attributes and preferences.

\paragraph{Implementation.} The adaptation ships as a separate module behind an environment flag; switched off, the LoCoMo protocol returns, serving as an ablation control. Both prompt sets appear in the supplementary material.

\section{Per-dialogue Results and Block-level Statistics}
\label{app:stats}

The 975 LoCoMo test questions nest within 5 dialogues sharing personas, events, banks, and retrieval errors---not independent samples; the main text's statistics therefore treat the dialogue as the independent unit (\S\ref{sec:setup}).

\begin{table}[h]
\caption{Per-dialogue accuracy (LoCoMo test split, 5 dialogues; 3-seed means).}
\label{tab:perdialogue}
\begin{center}
\small
\begin{tabular}{lrrrrrr}
\toprule
System & D6 & D7 & D8 & D9 & D10 & Mean \\
\midrule
Mem0 (shared top-10) & 68.0 & 71.2 & 69.5 & 70.4 & 69.4 & 69.7 \\
Mem0 (official $k{=}200$) & 74.0 & 75.6 & 74.3 & 75.2 & 74.4 & 74.7 \\
SFT (distillation start) & 61.8 & 63.9 & 62.1 & 62.6 & 62.1 & 62.5 \\
Outcome-only & 62.8 & 64.6 & 63.1 & 63.3 & 63.2 & 63.4 \\
Observational attribution & 68.4 & 71.6 & 69.9 & 70.7 & 70.4 & 70.2 \\
\textbf{Ours} & \textbf{75.2} & \textbf{79.1} & \textbf{76.8} & \textbf{78.3} & \textbf{78.1} & \textbf{77.5} \\
\bottomrule
\end{tabular}
\end{center}
\end{table}

\begin{figure}[h]
\begin{center}
\includegraphics[width=\linewidth]{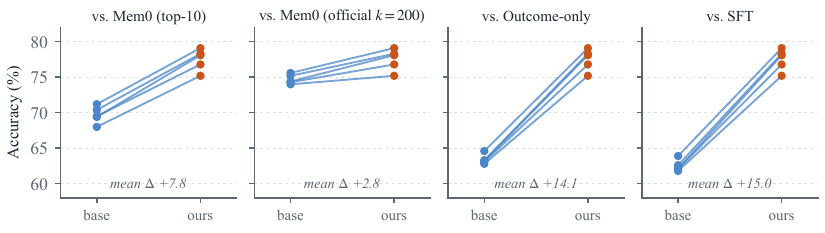}
\end{center}
\vspace{-6pt}
\caption{Per-dialogue paired slopes on the LoCoMo test split (D6--D10, 3-seed means; data of Table~\ref{tab:perdialogue}). Each line is one test dialogue; every pairing rises toward ours---the visual form of the 5/5 sign-test wins and of the recomputable block intervals.}
\label{fig:perdialogue}
\end{figure}

Our method exceeds Mem0's official $k{=}200$ configuration on all 5 dialogues. The pre-designated confirmatory comparison uses a one-sided paired sign test, giving $p = 1/32 \approx 0.031$, the smallest attainable value with 5 units; enumerating all $2^5$ sign flips gives the same permutation $p$-value. All other contrasts are descriptive and are not adjusted for multiplicity. Because only five dialogue blocks are available, the percentile bootstrap intervals below should also be interpreted descriptively rather than as the primary inferential result. The bootstrap ($10^4$ resamples, seed 0) gives a 95\% interval of $[76.1, 78.6]$ for our mean, and leave-one-dialogue-out means range from 77.1 to 78.1. All baseline and ablation arms have 3-seed standard deviations between $\pm$0.5 and $\pm$1.1; the $\pm$ entries in Table~\ref{tab:main} describe seed variability rather than dialogue-level uncertainty. Paired mean differences are 7.3 points vs.\ observational attribution (bootstrap interval $[7.0, 7.6]$), 7.8 vs.\ Mem0 top-10 ($[7.4, 8.3]$), 2.8 vs.\ Mem0 official $k{=}200$ ($[1.9, 3.5]$), 14.1 vs.\ outcome-only ($[13.2, 14.9]$), and 15.0 vs.\ SFT ($[14.1, 15.7]$). Our method is higher on all five dialogues in each comparison; Figure~\ref{fig:perdialogue} visualizes these paired differences.

\begin{table}[h]
\caption{Rollouts to reach gain thresholds (LoCoMo development set; both arms start from SFT at 63.1; thresholds are percentages of the gain to our plateau 78.0).}
\label{tab:sample}
\begin{center}
\small
\begin{tabular}{lrr}
\toprule
Threshold (absolute acc) & Outcome-only (within 28{,}000) & Ours \\
\midrule
Outcome-only plateau (64.0) & 9{,}600 & 120 \\
50\% of gain (70.6) & not reached (plateau 64.0) & 430 \\
75\% of gain (74.3) & not reached & 1{,}140 \\
90\% of gain (76.5) & not reached & 1{,}540 \\
Final plateau (dev) & 64.0 & 78.0 \\
\bottomrule
\end{tabular}
\end{center}
\end{table}

\begin{figure}[h]
\begin{center}
\includegraphics[width=\linewidth]{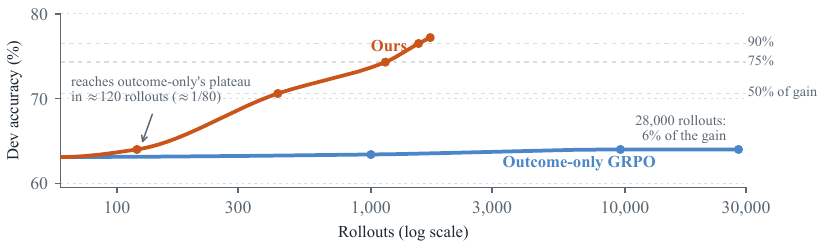}
\end{center}
\vspace{-4pt}
\caption{Learning curves (LoCoMo dev, log-scale rollouts); dotted lines mark 50/75/90\% of the gain. Exact numbers in Appendix~\ref{app:stats}.}
\label{fig:sample}
\end{figure}

\paragraph{Call-aligned and end-to-end accounting.} Our method uses approximately 1.6$\times$ as many API calls per rollout as outcome-only training because of the additional counterfactual re-answers. At the shared threshold of 64.0 accuracy, the rollout counts give $9{,}600 / (120 \times 1.6) \approx 50$, or an approximately 1:50 call-aligned ratio. Adding the one-time offline bootstrap (approximately 0.42M calls for teacher trajectories, probe generation, gating, and critic training) makes the two pipelines nearly equal at this early threshold (approximately 1:1.1). Over a complete run, the per-seed totals are approximately 0.55M calls for our method and 1.34M for outcome-only training, a 2.4$\times$ difference, while our final accuracy is 14 points higher. The larger 1:50 ratio therefore describes online efficiency at the shared threshold rather than total experimental cost.

On LongMemEval, the 400 questions are independent: ours vs.\ Mem0 per-question paired difference 15.0 points, paired bootstrap 95\% CI $[11.2, 18.8]$, McNemar $p < 10^{-6}$; by type, gains concentrate in multi-session reasoning (+19) and knowledge update (+21), with temporal +13, single-session +11, abstention +8---positive in all five.

\section{Control Matrices for Chain Credit and Retrieval Scaffold}
\label{app:chain}

\paragraph{Attribution controls for chain credit.} Eq.~(\ref{eq:credit})'s credit is shared un-normalized along version chains, admitting a mechanical explanation: the multi-version hubs of \S\ref{sec:emergent} could be artifacts of credit broadcasting rather than retrieval benefit, since merging into a high-value entry enlarges the rewarded set. The controls below test this, all retrained on LoCoMo with 3 seeds.

\begin{figure}[h]
\begin{center}
\includegraphics[width=0.9\linewidth]{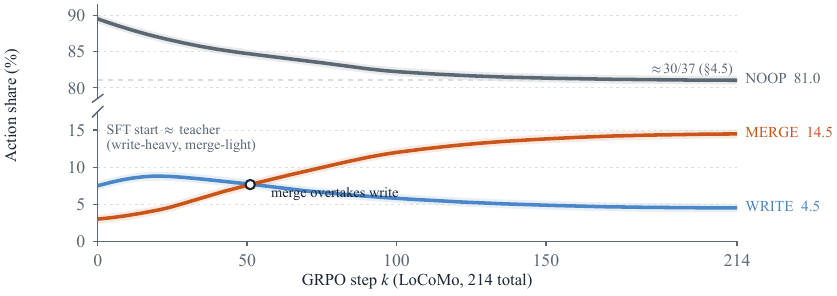}
\end{center}
\vspace{-6pt}
\caption{Action-share evolution over LoCoMo training (3-seed mean, $\pm$ band). From the SFT start (teacher-like: write-heavy, merge-light) the merge share rises fivefold ($3.0\%\to14.5\%$) and overtakes write by step ${\approx}50$, while write halves and \NOOP{} settles at the $\approx$30/37 level of \S\ref{sec:ablations}: the policy \emph{reallocates} actions toward consolidation rather than writing more, the behavioral trace of the organization in \S\ref{sec:emergent}.}
\label{fig:actions}
\end{figure}

\begin{table}[h]
\caption{Controls on the credit-sharing rule (LoCoMo, 975 questions).}
\label{tab:chaincontrols}
\begin{center}
\scriptsize\setlength{\tabcolsep}{4pt}
\begin{tabular}{lrrl}
\toprule
Configuration & Acc (\%) & Versions/entry & Note \\
\midrule
Main (shared along chain, un-normalized) & 77.5 $\pm$ 0.8 & 6.1 & \\
Normalized by chain length ($u_i/L$ per op) & 77.0 $\pm$ 0.9 & 5.4 & hubs persist, accuracy unchanged \\
Cited-version-only scoring (rest score 0) & 76.6 $\pm$ 1.0 & 5.2 & hubs persist \\
Chain sharing disabled (last contributor only) & 75.4 $\pm$ 1.1 & 3.9 & merge rate and accuracy both drop \\
\bottomrule
\end{tabular}
\end{center}
\end{table}

A per-version deletion spot check on 200 multi-version entries shows that cited versions account for 84\% of answer flips, so hub value is concentrated in cited versions rather than uniformly distributed with chain length. Multi-version organization persists under all three controls, and normalization retains similar structure and accuracy. These results make credit broadcasting unlikely to be the sole cause of the hubs. Disabling sharing lowers both merge rate and accuracy, suggesting that chain sharing encourages useful consolidation despite the overestimation bias discussed in \S\ref{sec:online} and \S\ref{sec:limitations}. \textbf{Choice of the un-normalized configuration.} The normalized configuration reaches 77.0 with smaller action overestimation and is a reasonable alternative. We retain the un-normalized form because the normalized arm requires approximately 40\% more steps to reach a stable merge rate. For future work, we recommend normalization when this additional optimization cost is acceptable.

\paragraph{Equalization controls for the retrieval scaffold.} The main-table comparison with external systems is system-level (Table~\ref{tab:main} caption). To isolate the retrieval interface's contribution:

\begin{table}[h]
\caption{Scaffold equalization controls (LoCoMo, 975 questions).}
\label{tab:scaffold}
\begin{center}
\scriptsize
\begin{tabular}{lr}
\toprule
Configuration & Acc (\%) \\
\midrule
Mem0 (official pipeline, shared top-10 reading) & 69.7 \\
Mem0 + our automatic verbatim-layer fallback & 70.9 \\
Heuristic teacher + explicit multi-version merge prompt & 65.9 \\
Rule-based merging (similarity clustering, same interface) & 66.5 \\
Heuristic teacher + usage-log feedback (two prompt iterations) & 67.4 \\
Ours (hit-version-only return, fixed token budget) & 75.6 \\
Ours $-$ automatic verbatim layer & 75.8 \\
Ours $-$ per-version embeddings (single entry embedding) & 74.6 \\
\textbf{Ours (full)} & \textbf{77.5} \\
\bottomrule
\end{tabular}
\end{center}
\end{table}

The verbatim layer and per-version retrieval interface each contribute approximately 2--3 points in their respective ablations, which is not enough to explain the 7.8-point gap to the strongest external system. Rule-based merging under the same multi-version interface reaches a similar version count (5.7) but only 66.5\% accuracy, showing that a high merge rate alone is insufficient. Two rounds of usage-log prompt feedback raise the teacher from 65.1 to 67.4, a 2.3-point improvement that does not close the gap. Replacing whole-chain return with hit-version-only return at a fixed token budget gives 75.6, indicating an interface contribution of about 1.9 points. Together, these controls suggest that the closed-loop signal affects which facts are consolidated, not only how often merges occur.

\paragraph{Sampled validation by action-level replay.} Entry-level presence credit and the marginal contribution of ``had this action not been taken'' are not the same quantity (\S\ref{sec:online}). We quantify the discrepancy on 300 stratified random operations (180 writes and 120 merges) by replaying the continuation after replacing the selected action with \NOOP{} under the same seed and decoding configuration. The final-exam outcome difference defines the replay marginal. Entry-level credit has Spearman correlation 0.72 with these marginals (bootstrap 95\% CI $[0.66, 0.77]$); the correlation is 0.78 $[0.71, 0.84]$ for writes and 0.61 $[0.50, 0.70]$ for merges. Divergence is concentrated in early writes later compensated by merges (12\%, credit overestimated) and precursor writes that provide attachment points for later merges (7\%, underestimated). Substituting replay marginals when recomputing the group advantages flips the sign for 8.7\% of the sampled operations, 83\% of which are merges; the resulting sign-flip-weighted gradient perturbation is below 5\% of the total gradient norm. This sample supports entry-level credit as an approximate action-supervision signal while quantifying its main biases. Replay is used only for validation because it requires one full continuation per operation.

\begin{figure}[h]
\begin{center}
\includegraphics[width=0.72\linewidth]{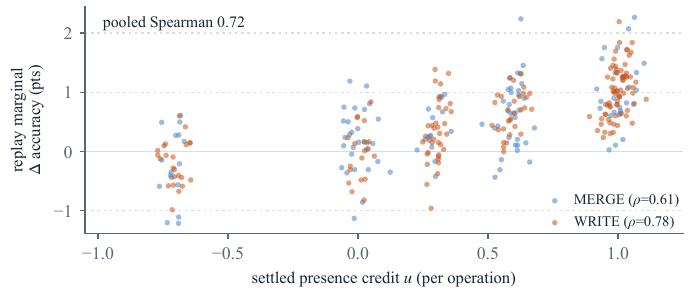}
\end{center}
\vspace{-6pt}
\caption{Action-level replay audit: settled presence credit vs.\ do($a\to$\NOOP{}) replay marginals over the 300 stratified operations. Pooled Spearman 0.72 (\WRITE{} 0.78, \MERGE{} 0.61), the correlations reported above.}
\label{fig:replay}
\end{figure}

\paragraph{End-to-end check with a different-family reader.} The main pipeline uses one API model for answering and judging (\S\ref{sec:setup}). To test whether the reported ordering depends on that reader, we re-evaluate four representative arms with GLM-5-Chat (glm-5-chat-20260701, temperature 0) as answerer and judge over 3 re-evaluation seeds. The banks are unchanged, but retrieval, answering, and judging are rerun. Accuracy is 74.8 $\pm$ 0.4 for our method, 67.9 $\pm$ 0.5 for Mem0, 61.2 $\pm$ 0.4 for outcome-only, and 60.4 $\pm$ 0.5 for SFT. Absolute values decrease by 2--3 points, but the ordering persists; our margin over Mem0 narrows from 7.8 to 6.9 points and remains positive on all 5 test dialogues.

\end{document}